\documentclass{article}

\PassOptionsToPackage{numbers,compress}{natbib}
\usepackage[preprint]{neurips_2026}

\usepackage[utf8]{inputenc}
\usepackage[T1]{fontenc}
\usepackage{hyperref}
\usepackage{url}
\usepackage{booktabs}
\usepackage{amsfonts}
\usepackage{nicefrac}
\usepackage{microtype}
\usepackage{xcolor}
\usepackage{graphicx}
\usepackage{flafter} % keep floats after their declaration
\usepackage{subcaption}
\usepackage{tabularx}
\usepackage{placeins}
\usepackage{colortbl}
\usepackage{makecell}
\usepackage{multirow}
\usepackage{paralist}
\usepackage{pifont}
\usepackage{amsmath}
\usepackage{amssymb}
\usepackage{mathtools}
\usepackage{amsthm}
\usepackage{adjustbox}
\usepackage{tikz}
\usepackage[capitalize,noabbrev]{cleveref}
\usepackage[disable,textsize=tiny]{todonotes}

\usetikzlibrary{positioning}
\tikzset{
  box/.style={draw, rounded corners, align=center, inner sep=6pt},
  inputbox/.style={draw, rounded corners, align=center, inner sep=6pt}
}

\newcommand{\gcheck}{\textcolor{green!60!black}{\ding{51}}}
\newcommand{\redx}{\textcolor{red!80!black}{\ding{55}}}

\theoremstyle{plain}

\theoremstyle{definition}

\theoremstyle{remark}

\newcommand{\methodname}{ORA}
\newcommand{\cumc}{CUMC}

\title{One Loss to Rule Them All: Marked Time-to-Event for Structured EHR Foundation Models}

\author{%
\begin{tabular}{c}
\textbf{Zilin Jing}$^{1,2,*}$ \quad \quad\quad
\textbf{Vincent Jeanselme}$^{2}$ \quad \quad\quad
\textbf{Yuta Kobayashi}$^{2}$ \\
\\
\textbf{Simon A. Lee}$^{3}$ \quad \quad \quad
\textbf{Chao Pang}$^{2,4}$ \quad \quad \quad
\textbf{Aparajita Kashyap}$^{2}$ \\
\\
\textbf{Yanwei Li}$^{2}$ \quad \quad \quad
\textbf{Xinzhuo Jiang}$^{2}$ \quad \quad \quad
\textbf{Shalmali Joshi}$^{2}$ \\[0.8em]
\normalfont $^{1}$Department of Computer Science, Columbia University \\
\normalfont $^{2}$Department of Biomedical Informatics, Columbia University \\
\normalfont $^{3}$Department of Computational Medicine, UCLA \\
\normalfont $^{4}$Formation Bio \\[0.4em]
\end{tabular}
}

\begin{document}

\maketitle

\begingroup
\begingroup
\renewcommand{\thefootnote}{}
\footnotetext{\textsuperscript{*}Corresponding author: \texttt{zj2398@columbia.edu}.}
\endgroup

% \renewcommand{\thefootnote}{}
% \footnotetext{\textsuperscript{\dagger}Corresponding author: \texttt{zj2398@columbia.edu}.}
% \endgroup

\begin{abstract}
Clinical events captured in Electronic Health Records (EHR) are irregularly sampled and may consist of a mixture of discrete events and numerical measurements, such as laboratory values or treatment dosages. The sequential nature of EHR, analogous to natural language, has motivated the use of next-token prediction to train prior EHR Foundation Models (FMs) over events. However, this training fails to capture the full structure of EHR. When a given event occurs must be captured, but the event value (abnormal lab)  also modulates the likelihood of other clinical events. Most existing EHR FMs do not jointly model this likelihood and are unable to capture the full observation process, impacting downstream capabilities. We propose \textit{\methodname},
% \footnote{Anonymized code available at \url{https://anonymous.4open.science/r/EHR_TTE/}}, 
% % \footnote{Codes will be released after }
a marked time-to-event pretraining objective that jointly models event timing and associated measurements. Across multiple datasets, downstream tasks, and model backbones, this objective consistently yields more generalizable representations than next-token prediction and pretraining losses that ignore continuous measurements. Importantly, the proposed objective yields improvements beyond traditional classification evaluation, including better regression and time-to-event prediction. 
Beyond introducing a new family of FMs, our ablations suggest a broader takeaway: pretraining objectives that account for EHR structure are critical for expanding downstream capabilities and generalizability.
\end{abstract}

\section{Introduction}
\begin{figure}[!htbp]
    \centering\includegraphics[width=0.9\linewidth]{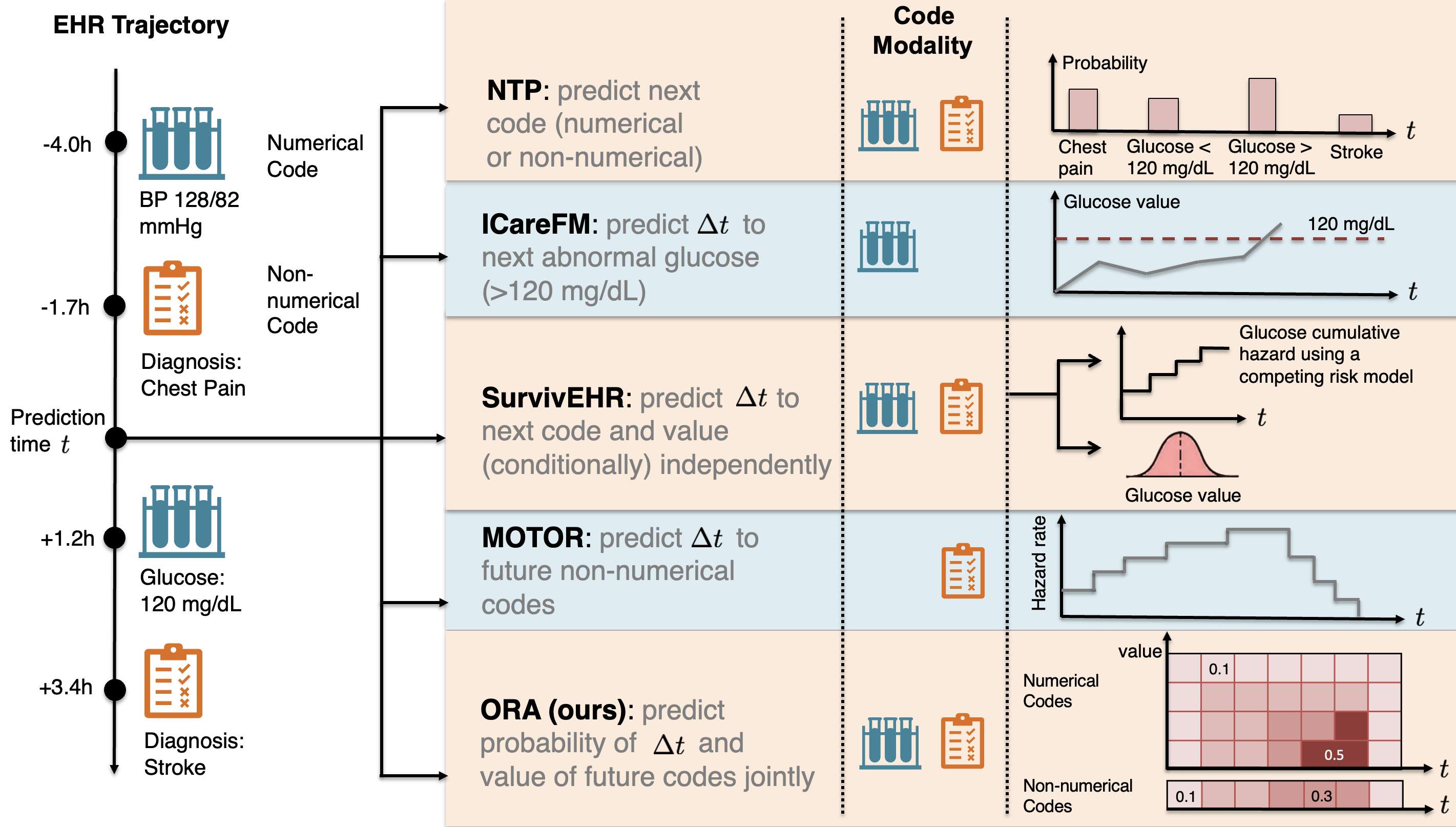} % TO UPDATE: https://docs.google.com/presentation/d/1pryqOkcbuJM58NkMXcq8wGuPa3oRy12-ZIEJ6uEeoh4/edit?usp=sharing
    \caption{\textbf{\methodname: Joint Modeling of Event Time and Value in EHR Pretraining.}
Comparison of conventional EHR pretraining objectives with ORA, illustrating how prior methods separately or partially model event sequences, whereas ORA jointly captures the distribution over future event timing and associated numerical values within irregular, heterogeneous clinical trajectories.}
    \label{fig:summary}
\end{figure}

Large-scale FMs have shifted the paradigm of machine learning in health by extracting representations that transfer across tasks, cohorts, and institutions~\citep{guo2023ehr, guo2024multi, lee2025clinical, burkhart2025foundation}. Inspired by the success of large language models (LLMs), recent work has framed Electronic Health Records (EHR) as an autoregressive sequence of discrete events pretrained using next-token prediction analogous to those used in natural language processing. These approaches have demonstrated state-of-the-art performance on a range of downstream classification tasks~\citep{li2020behrt,pang2024cehr,pmlr-v158-pang21a,odgaard2024core}.

Despite these advances, EHR data fundamentally differ from natural language in ways that challenge this analogy, motivating subsequent work. Clinical records are not sequences of tokens. Instead, they consist of irregularly sampled heterogeneous medical events, often accompanied by numerical measurements. To account for these deviations, most FMs design customized tokenizations, including artificial time tokens and time embeddings~\citep{pang2024cehr} for irregularity, discretized intervals~\citep{wornow2024context}, and digit space embedding~\citep{hur2023genhpf} for numerical value. However, these modifications focus on how data should be represented as input and often overlook the design of EHR pretraining losses. 

Recent literature~\citep{steinberg2024motor, gadd2025survivehr, burger2025foundation} proposed time-to-event pretraining loss as an alternative, with \citet{gadd2025survivehr, burger2025foundation} additionally introducing pretraining objectives that also capture numerical values. However, these works often introduce novel architectural choices, preprocessing, and tokenization strategies in addition to new pretraining losses. As a result, the impact of each design choice has been challenging to isolate. Further, when prior works model time-to-event (TTE) and mixed-event types, they either assume conditional independence between the time and value or model only abnormal values. A key missing element is the ability to jointly model time-to-event and mixed-event types (i.e., model values, if available, and marginalize them if not). We motivate this as a feasible tradeoff to a full joint likelihood of time, event-type, and values. %Our work aims to answer: \textit{Does a loss function accounting for EHR irregularity and numerical values yield more generalizable representations than the existing approaches?}

We propose \methodname, a marked time-to-event pretraining objective that jointly models the distribution of event timing and associated measurements for each medical code, naturally handling irregular intervals, censoring, and associated numerical values, if present. For the first time, we expand downstream evaluation of EHR FMs to characterize time-to-event and regression performance on multiple clinically meaningful tasks~\citep{pang2025fomoh}, to expand the clinical utility of EHR FMs. Across two large EHR datasets, representing large tertiary and quaternary healthcare institutions, we extensively evaluate the pretraining objective across $15$ tasks. We contribute a family of FMs, demonstrating that our proposed loss can be integrated with Transformer and Mamba backbones with improved performance on average across $7$ classification, $4$ time-to-event, and $4$ regression tasks across two different EHR datasets. Together, these results provide a critical insight into the design of future FMs: pretraining objectives should align with the full structure of EHR data to improve expressiveness, generalizability, and downstream capability.
Our contributions can be summarized as follows:
\begin{asparaitem}
    \item \textbf{Novel pre-training loss.} Building on the parallel between EHR and marked point processes, we introduce \methodname, a composite code-specific likelihood to capture the complexity of EHR models.
    %, effectively contributing a family of EHR foundation models.
    
    \item \textbf{Comprehensive evaluation and ablations.} Our evaluation is comprehensive: 7 binary classification, 4 TTE prediction, and 4 regression tasks across two datasets (MIMIC-IV~\citep{johnson2023mimic} and Columbia University Irving Medical Center (CUMC)). Our ablations demonstrate the crucial impact of the choice of pre-training loss for structured EHR FMs.

    \item \textbf{Generalizability.} We demonstrate improved generalizability of \methodname\, using both Mamba and Transformer base models (Section~\ref{sec:experiments}), across all tasks and dataets. 
\end{asparaitem}

\section{Background and Related Work}
\label{sec:literature}
This section reviews FMs for structured EHR data, with an emphasis on choices of pretraining objectives. We start by formulating the EHR representation learning problem as fitting a joint likelihood over irregularly sampled heterogeneous mixed-type events. 

{\bf{Formulation}}
Consider each patient $i \in [1, 2, \cdots, N]$'s EHR data to consist of a tuple $\mathcal{H}_i:= \{(t,m,v)_{i,j}, j \in [1, 2, \cdots, N_i]\}$ where $t \in \mathbb{R}^+$, is the timestamp of the event, $m \in \mathcal{M}$ the clinical code associated with the event, such as ICD-10 diagnosis or procedure codes, and $v \in \mathbb{R} \cup\{\varnothing\}$ the optional numerical value associated with the event. $N_i$ denotes the number of events observed for patient $i$. For a patient $i$, events up to $j$ (ordered by time) are denoted by  $\mathcal{H}_{i, < j}$. That is, $\mathcal{H}_{i,< j} := \{(t,m,v)_{i,l}, l \in [1, 2, \cdots, j-1]\}$. An intuitive approach is to maximize the joint likelihood of the full marked point process, corresponding to the joint between the time to the next event, the associated code, and value. Under standard IID assumptions on patient samples, the joint likelihood can be decomposed as: 
\begin{align}\label{eq:joint_likelihood}
\mathcal{L}(\theta)
= \prod_{i=1}^{N} p_\theta(\mathcal{H}_i)
= \prod_{i=1}^{N}\prod_{j \in [1..N_i]} p_\theta((t,m,v)_{i,j} \mid \mathcal{H}_{i, < j})
\end{align}

In the following sections, we discuss three different types of pretrained loss functions from prior work and contextualize how they approximate the joint likelihood. 

\textbf{Next Token Prediction (NTP)}: 
Inherited from natural language, the NTP loss collects all medical codes $m$ as a vocabulary and predicts the next token in the trajectory autoregressively. This is equal to maximizing a simplified likelihood: \[
\mathcal{L}_{\text{NTP}}(\theta) := \prod_{i=1}^{N}\prod_{j \in [1..N_i]} p_\theta(m_{i,j} \mid \mathcal{H}_{i, <j})
\]

% Even in the more general context of time series, FMs often minimize this loss~\citep{ansari2024chronos} or the mean square error of the next value~\citep{jin2023time, goswami2024moment, chang2025llm4ts, das2024decoder}, implicitly assuming temporal regularity in observation sequences. 

However, such losses do not account for the temporal irregularities and associated values characteristic of EHR, a problem that has largely been overlooked in the development of FMs. To overcome these limitations, several NTP models design EHR-specific tokenization strategies: (i) train separate embedding layers for time and value, synthesized with code embeddings as input \cite{fallahpour2024ehrmamba,pang2021cehr}. Such tokenizations do not treat elapsed time and value as an explicit supervised target, making it hard for embeddings to capture an irregular data-generating process, (ii) adding discretized tokens for time and value to the vocabulary \cite{hegselmann2025large,pang2024cehr,renc2024zero}. The cross-entropy loss is computed for all $t$, $m$, and $v$ tokens. For example, Context Clue \cite{wornow2024context} discretizes each numerical measurement to ten intervals and assigns each interval a unique token during pretraining. However, this method blows up the vocabulary size in order to get a precise division of each numerical code. Meanwhile, it ignores censoring in the loss function, which causes structural missingness and incomplete EHR trajectories \cite{al2024biases}.

\textbf{Temporal Point Process (TPP)}:
To capture the underlying irregular data-generating process, TPP directly predicts the time interval to the next event and its code type \cite{shmatko2025learning}:
    \[
\mathcal{L}_{\text{TPP}}(\theta) :=\prod_{i=1}^N\prod_{j}p_{\theta}(\Delta t_{i,j}, m_{i,j} \mid \mathcal{H}_{i,< j})
\]

The probability is computed by integrating the intensity function $\lambda(t)$ on each time interval. However, this parametrization is sensitive to simultaneous events where the inter-event interval $\Delta t \approx 0$, causing the intensity function to become extremely unstable during pretraining. To address this, MOTOR~\citep{steinberg2024motor} separately models the next event of each code and computes the mean likelihood over all codes. Both methods only model non-numeric codes and ignore the value associated with numerical events.

 %Particularly, FMs' training often relies on maximizing the likelihood of the next token.  

{\bf{Marked Temporal Point Process (MTPP).}} 
Two most recent EHR Foundation models approximate the joint likelihood with the marked temporal point process \cite{gadd2025survivehr} \cite{burger2025foundation}. SurvivEHR \cite{gadd2025survivehr} uses competing risk to model the inter-event time and fit a code-specific Gaussian distribution for numerical codes with the following likelihood:
    \[
\mathcal{L}_{\text{SurvivEHR}}(\theta) :=\prod_{i=1}^N\prod_{j}p_{\theta_1}(\Delta t_{i,j} \mid m_{i,j}, \mathcal{H}_{i,< j})p_{\theta_2}(v_{i,j} \mid m_{i,j}, \mathcal{H}_{i,< j})
\]
Similar to TPP, this method struggles with concurrent events. Additionally, it assumes conditional independence between time and value, thereby biasing the joint likelihood. ICareFM \cite{burger2025foundation} addresses these limitations by defining a surrogate loss function for a set of abnormal events $K$ where the numerical codes have exceptionally low or high values, equivalent to maximizing the joint likelihood of all abnormal events:
    \[
\mathcal{L}_{\text{ICareFM}}(\theta) :=\prod_{i=1}^N\prod_{(m_{i,j},v_{i,j})\in K}p_{\theta}(\Delta t_{i,j} \mid \mathcal{H}_{i,< j})
\]
This method is exclusive to numerical measurement and ignores other nonnumerical codes like diagnosis and procedures.

Modeling the complete joint likelihood in Equation \eqref{eq:joint_likelihood} is challenging since we need to capture the complex relationship between time, code, and value. EHR events also co-occur frequently, making it hard to fit the intensity function for a standard point process. 

% To addresses these challenges, we instead propose a new loss function and a family of marked time-to-event foundation models in section~\ref{sec:method}. Additionally, all previous FMs have  various tokenization strategies and architectural differences. To isolate the impact of the choice of pretraining loss on downstream performance, we carry detailed ablation studies in section~ \ref{sec:experiments} by fixing the tokenizer and the architecture.

\section{Method: \methodname}
\label{sec:method}

To address these challenges, we instead propose optimizing the composite code-specific likelihood. We denote by $f_{i, j}^m$, the first occurrence of code $m$ after the time associated with event $j$, and the set of the first events of all codes as $\mathcal{F}_{i,j} = \{\forall m, f_{i,j}^m\}$.  Specifically, we define:
\[
f_{i, j}^m = (\Delta t_j^m, v, \delta, m)
\] 
where, if an event of type $m$ occurs after $t_j$, 
$\Delta t_j^m$ corresponds to the interval between the current time $t_j$ and the time to the next event $m$, $v$ is the associated numerical value with the future event, and $\delta = 1$ as an event indicator. If no such event occurs in the future, we denote $\Delta t_j^m = t_{i,N_i} - t_{i,j}$ as the interval up to the last observation, and set the event indicator to $\delta = 0$. In this context, $\delta$ denotes whether an event $m$ is observed within the recorded EHR for patient $i$ after the observation $j$ or censored. Accounting for such unobserved events is critical, as ignoring them leads to biased likelihood estimates~\citep{chen2024introduction}. Suppose all the first occurrences of future codes are conditionally independent given the contextual history $\mathcal{H}_{i,< j}$ of the patient:
\begin{align}
\Tilde{\mathcal{L}}_{\textsc{ORA}}(\theta)
&= \prod_{i}\prod_{j} \prod_{(\Delta t, m, v, \delta) \in \mathcal{F}_{i, j}} p_{\theta_m}(\Delta t, v, \delta \mid \mathcal{H}_{i,< j})
\label{eq:mtte_joint_likelihood}
\end{align}
Formally, this corresponds to a code-specific marked point process. This formulation predicts the joint probability of time and value, capturing the complex dependence between different properties within a medical event. Meanwhile, it empirically allows the use of all potential next events at a given time point, thereby overcoming the sparsity of observations and improving data efficiency. Finally, \methodname~is robust for both nonnumerical and numerical events. If $v$ is missing or empty, we can marginalize the probability over $v$. The detailed comparison between \methodname~and existing EHR FMs losses can be found in Table~\ref{fig:comparison}.

\subsection{Discretized \methodname}
% Express without approximation
The central challenge in computing the proposed loss in Equation~\eqref{eq:mtte_joint_likelihood} is the estimation of $p_{\theta_m}(\Delta t, v, \delta \mid \mathcal{H}_{i,< j})$, or equivalently, under the conditional independence and the common uninformative censoring assumptions:
\begin{align*}
&\log p_{\theta_m}(\Delta t, v, \delta \mid \mathcal{H}_{< j})
= \delta\log \lambda_{\theta_m}(t, v \mid \mathcal{H}_{< j}) - 
\int_{0}^{t} \int_{v'} \lambda_{\theta_m}(u, v' \mid \mathcal{H}_{< j}) dv' dt'
\end{align*}
where $\lambda_{\theta_m}(\cdot)$ is the instantaneous risk, or the intensity function, of observing an event code $m$.

% Describe challenges in estimating
Estimating the likelihood, therefore, requires modeling the instantaneous risk and its integral for each code $m$. Approaches in survival analysis and temporal point process analysis often constrain the intensity function to have a closed-form integral through parametric assumptions~\citep{du2016recurrent, mei2017neural} or through constrained neural networks~\citep{omi2019fully, jeanselme2023neural}. In this work, we adopt a discretization of the joint, extending survival approaches such as DeepHit~\citep{lee2018deephit} to marked point processes. While approximating the likelihood, this approach has been shown to be computationally efficient and to achieve state-of-the-art performance in survival analysis~\citep{lee2018deephit}. Concretely, we add a final network layer to model the discretized joint distribution over time-value tuples. For each code $m$ and input $x$, the neural network outputs a matrix $P^m(x) \in [0, 1]^{T \times V}$, a discretized approximation of the intensity function, where $T$ and $V$ denote the number of discretized bins for time and value, respectively. Each entry $P^m[k,l]$ represents the probability of observing an event in the $k^{th}$ time-quantile and $l^{th}$ value-quantile. Note that these quantiles are code-specific to ensure more flexible modeling. Let $q_m(t, v)$ return the time-value quantiles associated with code $m$ to which $(t,v)$ belongs. When $v$ is omitted, $q_m(t, \cdot)$ denotes the time-quantile. Under this discretization, the log-likelihood can be expressed as:

% \begin{align*}
% \log \tilde{\mathcal{L}}_{\textsc{\methodname}}(\theta) := \sum_{i,j}\sum_{(t, m, v, \delta) \in \mathcal{F}_{i, j}} &\delta
% \log P_{\theta}^m[q_m(t, v)](\mathcal{H}_{i,< j})
% + (1-\delta)  \log\left(1- \sum_{k = 1}^{q_m(t, \cdot)} \sum_{l} P_{\theta}^m[k, l](\mathcal{H}_{i,< j}) \right)
% \end{align*}

\begin{align*}
\log \tilde{\mathcal{L}}_{\textsc{\methodname}}(\theta) := \sum_{i=1}^{N} \sum_{j \in [1...N_i]} \sum_{(t, m, v, \delta) \in \mathcal{F}_{i, j}} &\delta
\log P_{\theta}^m[q_m(t, v)](\mathcal{H}_{i,< j})\\ 
+ &(1-\delta)  \log\left(1- \sum_{k = 1}^{q_m(t, \cdot)} \sum_{l} P_{\theta}^m[k, l](\mathcal{H}_{i,< j}) \right)
\end{align*}
where $q_m((t, v))$ returns the quantile to which $(t, v)$ belongs for mark/event $m$. 
For observed events, this objective maximizes the probability mass assigned to the quantile containing the time and value of the next event, which is equivalent to a standard cross-entropy loss. For censored events, it enforces a low probability of observing any event before the end of the observation window, marginalizing over all possible values. The number of time and value bins, $T$ and $V$, are shared hyperparameters, while the corresponding quantile boundaries are computed separately for each code from the empirical distributions of inter-event times and observed values. Thus, different codes share the same output dimensionality for the probability matrix but have code-specific discretization intervals, allowing the bins to adapt to heterogeneous event frequencies and value ranges. Additional details and ablation results on the choice of $T$ and $V$ are provided in Appendix~\ref{sec:hyperparameter}.

% Both T and V are code-specific quantiles derived from the empirical distribution of inter-event time and the value associated with each code. This allows different codes to have customized resolutions of the discretized interval. Further details can be found in the Appendix~\ref{sec:hyperparameter} , where we report ablation studies results for investigating how different hyperparameters impact the downstream performance. 

\subsection{Model design}
To maximize the discretized \methodname~likelihood, one must extract a representation of the medical history $\mathcal{H}_{i, <j}$ and compute the matrix $P^m(x)$. Importantly, the proposed pre-training loss is backbone agnostic. We propose evaluating its efficacy on two common FMs backbones: the attention-based Transformer~\citep{vaswani2017attention}, and a state-space model (Mamba~\citep{gu2024mamba}) to demonstrate its utility. 

% These models are only altered in their output layers by adding XX to output the previously described matrix $P$.

{\bf{EHR tokenization.}}
Following \citet{steinberg2024motor}, we use a hierarchical tokenizer. For each event, we construct its embedding by summing up the embeddings of each associated attribute: e.g., code, numerical value, etc. This joint event encoding has been demonstrably better than factorizing the event into multiple tokens \cite{guo2026tokenization}. All numerical values are pooled together and discretized into shared quantiles. During tokenization, each continuous value is assigned a quantile token for its value embedding. For codes, we adopt an entropy-based filter to construct the vocabulary with the most informative events~\citep{steinberg2024motor}. We quantify the informativeness of each code using its empirical entropy, and retain codes with the highest entropy to reduce noise during learning (see details in Appendix~\ref{app:tokenizer}). 

{\bf{Backbone.}} Following \citet{steinberg2024motor}, and maintaining consistency to demonstrate the gains from the choice of pre-training loss, we adopt a decoder-only Transformer backbone that avoids look-ahead from future medical events through a causal attention mechanism. Unlike the standard Transformer, it uses rotary position embeddings with age for improved temporality processing. It also adopts local attention and sample packing for efficient pretraining. To demonstrate the generalizability of our pretraining loss, we also evaluate a Mamba-base, a state-space model \citep{gu2024mamba}, as an alternative backbone. Mamba replaces attention-based modules in the Transformer with a selective state-space block. Prior work has shown its advantage over the Transformer in capturing long, irregular EHR sequences with temporal dependencies \citep{wornow2024context, fallahpour2024ehrmamba}. The detailed model configurations can be found in Appendix~\ref{sec:model_configurations}.

\FloatBarrier
% {\bf{Efficient projection head.}}
% The last layer of our architectures includes a head to project the output embedding $E \in \mathbb{R}^D$, with $D$ is the model's hidden dimension, to the probability matrix $\forall m \in \mathcal{M}, P^m(x) \in (0, 1)^{T \times V}$. A direct projection from $\mathbb{R}^D$ to the discretized joint of shape $T \times V \times |\mathcal{M}|$ would require an impractical number of parameters to estimate. Instead, following \citep{steinberg2024motor}, we use a factorized two-stage computation.
% First, a one-layer fully connected neural network projects the embedding $E$ to time-specific features $H_j \in \mathbb{R}^{T \times D_2}$.
% Then we use a second projection followed by a final Softmax for all codes with numerical values, projecting $H_j$ into the final matrix $P^m \in [0,1]^{T \times V}$. Similarly, for nonnumerical codes, a single-layer project into $P^m \in [0,1]^{T \times 1}$.

% In our experiments, we use $D=768$, $D_2=512$, $T=8$, $V=10$, and $|\mathcal{M}|\approx 7000$. This factorization reduces the number of parameters by $20$\% compared to a fully connected network that projects the embedding onto the discretized joint. Full architectural details and exact tensor shapes are provided in Appendix~\ref{app:compression}. All models are trained with a fixed 120M-parameter budget; model configurations are listed in Appendix~\ref{app:hyperparam}.

\section{Experiments}
\label{sec:experiments}
Our experiments characterize the impact of the choice of pretraining loss on the representativeness of extracted embeddings. We compare prior EHR FMs to ORA using two base backbones and conduct linear probing to evaluate performance on $15$ tasks across two EHR datasets. Our ablations further explicitly compare NTP, TPP, and ORA to isolate the impact of the proposed pre-training loss.  

% #a private EHR dataset from a large urban hospital called `Institution' hereafter
\subsection{Datasets}
We rely on one public EHR dataset: MIMIC-IV~\citep{johnson2023mimic} and a private EHR dataset from Columbia University Irving Medical Center (CUMC). MIMIC-IV contains 364,000 patients who were admitted to Beth Israel Deaconess Medical Center between 2008 and 2022, with tabular time-series measurements of diagnoses, medications, lab tests, etc. The \cumc{} dataset contains 6.7 million patients with both inpatient and outpatient visits, primary care, and emergency services between $1986$ and $2023$, consisting of diagnoses, procedures, medications, and visit information (see Appendix~\ref{app:datasets} for patient split used for pretraining). Additional details about tokenization, architecture design, and hyperparameter tuning are in Appendix~\ref{app:model}.

\subsection{Downstream tasks}
To ensure clinical utility, we design a set of clinically meaningful downstream tasks that encompass binary classification, time-to-event prediction, and regression of lab tests. 

{\bf{Binary Classification.}}
We consider 7 binary classification tasks from~\citet{pang2025fomoh}:
\begin{asparaitem}
    \item Predicting three patient outcomes: In-hospital mortality (Mortality), a length of stay longer than $7$ days (LOS), and $30$-day readmission (Readmission).
    \item Predicting onset of two chronic conditions: Celiac and Metabolically Associated Steatotic Liver Disease (MASLD). 
    \item Predicting two acute conditions: Acute Myocardial Infarction (AMI) and Ischemic Stroke (Stroke). 
\end{asparaitem}

The detailed definition of phenotype diseases can be found in Appendix~\ref{app:classification}. Predicting chronic and acute conditions is particularly challenging because of the considered `at-risk' cohort, which increases task difficulty while improving downstream clinical utility~\citep{pang2025fomoh}. We treat chronic conditions as incident onset events and therefore evaluate only the first occurrence, consistent with clinical interest in early and first identification. In contrast, acute conditions may recur, so we include all occurrences and formulate a recursive prediction task to capture repeated risk over time.

{\bf{Time-to-event.}}
Clinical decision-making often depends not only on whether an event will occur, but also on when it is expected to occur. To evaluate this capability, we consider time-to-event prediction for four conditions (Celiac, MASLD, AMI, and Stroke). For each condition, the prediction time is defined at the discharge of an inpatient visit, and the task is to predict the time until the patient receives the first diagnosis of the corresponding condition. 

{\bf{Regression.}}
Comprehensive characterization of patient health requires predicting future physiological measurements. Prior works have largely overlooked regression and TTE as downstream tasks. We assess FM performance on regression tasks involving four important laboratory measurements for sepsis patients: Platelets, Creatinine, Oxygen, and Glucose. Each measurement is identified by a single medical code (see Appendix~\ref{app:regression}). All regression tasks are included in the tokenizers' vocabularies for each pretrained model, ensuring they are not out of vocabulary at evaluation time. To avoid leakage across visits, we restrict evaluation to same-visit laboratory tests that occur at least 4 hours after admission, with the prediction time as exactly 4 hours before the corresponding lab event.

\begin{figure}[t]
    \centering
    \includegraphics[width=\linewidth]{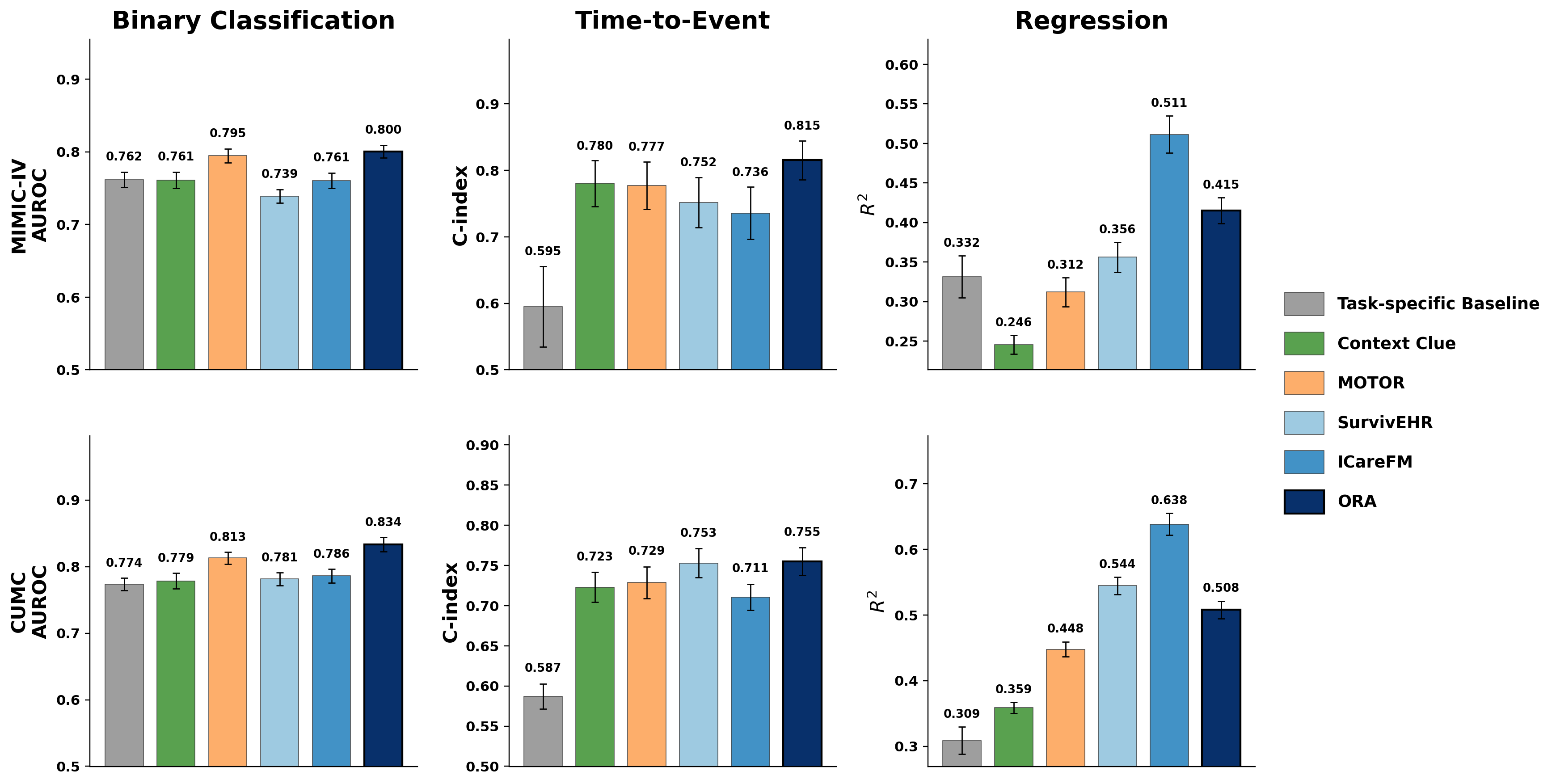}
\caption{\textbf{ORA Improves Generalization Across Classification, Time-to-Event, and Regression Tasks.} Average performance on MIMIC-IV and Columbia
University Irving Medical Center (CUMC) datasets shows that jointly modeling event timing and values yields consistently stronger representations. ORA outperforms  different foundation models across both binary classification and TTE tasks and remains competitive in regression results. All models used the \textbf{Transformer} backbone. Full results for each task can be found in Appendix~\ref{app:results}.
% on MIMIC-IV and \cumc~ datasets. All models used the \textbf{Transformer} architecture.
}
    \label{fig:mimic_baseline}
\end{figure}

% \begin{figure}[t]
%     \centering
%     \includegraphics[width=\linewidth]{Figure/mimic_ablation.png}
% \caption{Ablation studies of different loss function on each MIMIC-IV task for the \textbf{Transformer} architecture. We use the same tokenizer and the architecture for all models for a fair comparison.}
%     \label{fig:mimic_ablation}
% \end{figure}

\subsection{Baseline models and associated loss}
Our experiments include the following baselines and FMs:

\begin{asparaitem}
    \item \textbf{Task-Specific Baselines.} For each task, we consider XGBoost~\citep{chen2016xgboost} for binary classification, Deephit~\citep{lee2018deephit} for time-to-event prediction based on FEMR features~\citep{steinberg2024motor} from the EHR time series. For regression, we used the most recent value for the same lab code measured before the prediction time, which is a strong baseline that outperforms both structured EHR foundation models and general LLM (GPT-4o) in lab outcome predictions \cite{im2025labtop}. 
    \item \textbf{Context Clue} \cite{wornow2024context} is a  next-token prediction (NTP) EHR FM family. We include two different backbones: CC-Llama and CC-Mamba. 
    \item \textbf{Motor} \cite{steinberg2024motor} is a time-to-event (TTE) EHR foundation model pretrained to predict code-specific future event risks. It serves as an external baseline for temporal point process-style pretraining, which models event timing but does not explicitly model the associated numerical value distribution.
    \item \textbf{SurvivEHR} \cite{gadd2025survivehr} and \textbf{ICareFM} \cite{burger2025foundation} are two marked time-to-event foundation models (MTPPs) that jointly predict time and value in the pretraining losses.
    These baselines are closest in spirit to \methodname, while differing in their loss functions.
\end{asparaitem}

\subsection{Ablation studies}
To exclusively study the impact of loss function and understand how each component (medical code, time and numerical value) in our \methodname~loss contributes to the downstream performance, we fix the tokenizer and backbone of \methodname. We create three ablations with the loss functions as follows: 

\begin{asparaitem}
\item \textbf{NTP.}
We train standard next-token prediction baselines that autoregressively predict the code type of the next medical event using cross-entropy loss. This objective supervises the next code only, without directly modeling event time or numerical value.

\item \textbf{TPP.}
We train temporal point process-style baselines that predict, for each medical code, the time interval to its first future occurrence. This objective adds time-to-event supervision while marginalizing over the numerical values.

\item \textbf{\methodname.}
We train the full marked time-to-event objective proposed in this work. In addition to predicting the time interval as in TPP, \methodname~jointly models the value distribution for numerical codes such as laboratory tests and vital signs.
\end{asparaitem}

\subsection{Evaluation}
To quantify the discriminative performance of all models, we report the Area Under the Receiver Operating Characteristic Curve (AUROC) for classification, the time-dependent C-index~\citep{antolini2005time} for TTE, and $R^2$ coefficient for regression. %For comparison, we present the relative performance gain over the NTP baseline for the best method in each loss family. 
When evaluating the pretrained models, we generate feature embeddings for each task cohort and fit a task-specific head (logistic regression for binary classification, Deephit for time-to-event tasks, and linear regression for regression tasks).

% To further study the impact of our loss function, we reconstruct all baselines by using the same tokenizer and architecture as our ORA models. Therefore, the performance gap is entirely due to the different loss functions. Detailed model performance is provided in Appendix~\ref{app:results}.

\begin{figure}[t]
    \centering
    \includegraphics[width=\linewidth]{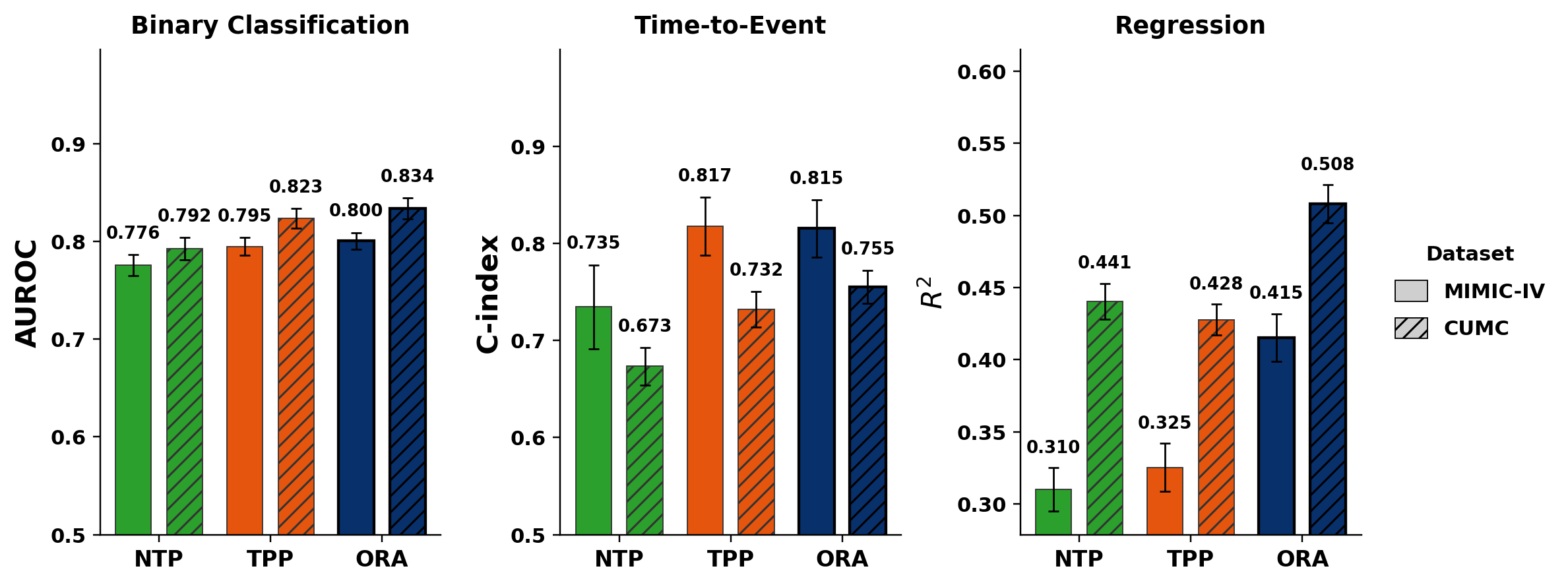}
\caption{\textbf{Pretraining Objective Drives Downstream Performance Across Task Families.}
Under a fixed tokenizer and Transformer backbone, jointly modeling time-to-event and values yields consistent gains over NTP and TPP objectives in most tasks, demonstrating that improvements arise from the loss formulation rather than model capacity or input representation.}
    \label{fig:mimic_ablation}
\end{figure}

\section{Results}
\label{sec:results}

\subsection{ORA improves overall task performance across datasets}

Figure~\ref{fig:mimic_baseline} compares ORA with task-specific baselines and representative EHR foundation models on top of the Transformer backbone. Across both MIMIC-IV and CUMC, \methodname~achieves the most balanced performance across binary classification, time-to-event prediction, and regression. On MIMIC-IV, \methodname~obtains the best average AUROC for binary classification and the best average C-index for time-to-event prediction, reaching $0.800$ and $0.815$, respectively. These results improve over the strongest non-\methodname~baselines, including MOTOR and Context Clue. On CUMC, the same trend holds: \methodname~achieves the best average binary classification performance ($0.834$ AUROC) and remains best or nearly tied for time-to-event prediction ($0.755$ C-index).

For regression tasks, \methodname~consistently outperforms NTP and TPP baselines. While specialized MTPP models such as SurvivEHR and ICareFM achieve the highest $R^2$ in regression tasks, this performance comes at the cost of severe degradation in classification and TTE. This trade-off is  expected since ICareFM relies on a surrogate loss function that isolates abnormal numerical events, effectively over-indexing on numeric data while ignoring non-numerical codes. SurvivEHR is sensitive to highly concurrent events, and also assumes the conditional independence between time and value. In contrast, \methodname~jointly models the distribution of event timing and associated measurements, achieving the best overall performance across three different types of tasks.

\subsection{Ablation studies: isolating the impact of the \methodname~loss function}

To isolate the impact of the pretraining objective, Figure~\ref{fig:mimic_ablation} compares different objectives under a fixed tokenizer and Transformer backbone. This ablation directly tests whether the gains of \methodname~come from its marked time-to-event objective rather than from architectural or tokenization differences. In this controlled setting, ORA consistently outperforms NTP across all three task families on both datasets. These gains show that predicting only the next discrete event is insufficient for learning general-purpose EHR representations, since it ignores both irregular timing and event-associated numerical values. 

Compared with TPP, which adds time-to-event supervision but does not jointly model numerical values, \methodname~preserves the temporal benefits of TTE modeling while substantially improving tasks that depend on measurement values. This comparison shows that time modeling alone is not sufficient: the additional value component in \methodname~is critical for regression and also improves classification without sacrificing time-to-event performance. Overall, the ablation demonstrates that the marked time-to-event loss itself is the key driver of the observed robustness across tasks and datasets.

\subsection{\methodname's performance generalizes across backbones}

Finally, we demonstrate that the representational gains of the \methodname~loss are fundamentally backbone-agnostic. We replaced the attention-based Transformer with Mamba, a state-space architecture. 
Figure~\ref{fig:mimic_baseline_mamba} in Appendix~\ref{sec:mamba_results} further supports this conclusion. Although MOTOR is slightly stronger for binary classification and ICareFM is the strongest for regression, \methodname~again achieves the most balanced performance across task families, with strong classification, the best time-to-event performance, and competitive regression. Meanwhile, Figure~\ref{fig:mimic_ablation_mamba} shows that moving from NTP or TPP to ORA loss is also beneficial for Mamba backbones.
Together with the Transformer results, these findings suggest that the advantage of \methodname~comes from the pretraining objective itself rather than a particular backbone. Across datasets and backbones, jointly modeling when clinical events occur and what values they take yields more general-purpose EHR representations than objectives that model only tokens, only time, or only numerical deterioration events.

\section{Discussion}
We introduce \methodname{}, a marked time-to-event pretraining objective for structured EHR foundation models. The central motivation of this work is that EHR trajectories are not simply sequences of discrete tokens: clinical events occur irregularly over time, may be observed concurrently, and are often associated with numerical values. Existing pretraining objectives only partially capture this structure. Next-token prediction treats EHR as an ordered sequence of discrete codes. Time-to-event objectives model event timing but often ignores the complex dependence between time and value, fails to capture mixed-type events and it is sensitive to concurrent events.

Our empirical results support the importance of aligning the pretraining objective with the structure of EHR data. Across two EHR datasets, three downstream task families, and two backbone architectures, \methodname{} produces representations that are consistently more generalizable than next-token prediction and time-to-event objectives that ignore numerical measurements. In particular, \methodname{} improves not only binary classification, the dominant evaluation setting in prior EHR foundation model work, but also time-to-event prediction and numerical regression. This broader evaluation highlights that EHR foundation models should be assessed beyond classification, since clinically useful representations must support heterogeneous prediction targets.

A key strength of our study is that we isolate the effect of the pretraining loss. By fixing the tokenizer and backbone in our ablation studies, we directly compare \methodname{} against next-token prediction and time-to-event objectives under controlled settings. These experiments show that jointly predicting event timing and value improves downstream performance overall, with especially strong gains on regression tasks. The same trend holds across both Transformer and Mamba backbones, suggesting that the benefit of \methodname{} is not tied to a specific sequence model. More broadly, our results indicate that pretraining objectives, rather than backbone alone, are a critical design choice for structured EHR foundation models.

{\bf{Limitations and future work.}} There are several limitations. First, \methodname{} relies on quantile-based discretization of time and value. This choice is simple, scalable, and adapts to code-specific distributions, but it may be suboptimal for sharply peaked or highly multimodal distributions. More expressive discretization schemes, adaptive binning, or continuous likelihood parameterizations could further improve the fidelity of the learned event-time and value distributions.

Second, our composite likelihood assumes that the first future occurrences of different codes are conditionally independent given the patient history. This assumption improves scalability and allows dense supervision over many possible future events, but it is still a simplification of the full joint EHR process. Future work could relax this assumption by explicitly modeling dependencies among future codes, while preserving the computational advantages of the current formulation.

Third, although \methodname{} is robust across datasets, tasks, and backbones, it does not uniformly achieve the best performance on every regression task. Some numerical outcomes may benefit from objectives that more directly model physiological value dynamics or abnormal-threshold events. Combining \methodname{} with complementary value-centric objectives, such as threshold-conditioned or continuous-value prediction losses, may further improve robustness on regression-heavy settings.

\begin{ack}
VJ, and SJ would like to acknowledge partial support from NIH 5R01MH137679-02. AK would like to acknowledge the support of NIH 5T15LM007079-34. ZJ and SJ would like to acknowledge partial support from NIH 5R01LM014380-02. SJ and ZJ would like to acknowledge partial support from the Google Research Scholar Award. SAL would like to acknowledge the generous support of the Warren Alpert Foundation, Warren Alpert-UCLA Computational Biology/AI (CBAI) Scholar Training and Retention Program (award 20242540). SJ would like to acknowledge partial support from the SNF Center for Precision Psychiatry \& Mental Health, Columbia. Any opinions, findings, conclusions, or recommendations in this manuscript are those of the authors and do not reflect the views, policies, endorsements, expressed or implied, of any aforementioned funding agencies/institutions.
\end{ack}

\newpage
\bibliographystyle{plainnat}
\bibliography{references}

\begin{thebibliography}{33}
\providecommand{\natexlab}[1]{#1}
\providecommand{\url}[1]{\texttt{#1}}
\expandafter\ifx\csname urlstyle\endcsname\relax
  \providecommand{\doi}[1]{doi: #1}\else
  \providecommand{\doi}{doi: \begingroup \urlstyle{rm}\Url}\fi

\bibitem[Al-Sahab et~al.(2024)Al-Sahab, Leviton, Loddenkemper, Paneth, and Zhang]{al2024biases}
Ban Al-Sahab, Alan Leviton, Tobias Loddenkemper, Nigel Paneth, and Bo~Zhang.
\newblock Biases in electronic health records data for generating real-world evidence: an overview.
\newblock \emph{Journal of healthcare informatics research}, 8\penalty0 (1):\penalty0 121--139, 2024.

\bibitem[Antolini et~al.(2005)Antolini, Boracchi, and Biganzoli]{antolini2005time}
Laura Antolini, Patrizia Boracchi, and Elia Biganzoli.
\newblock A time-dependent discrimination index for survival data.
\newblock \emph{Statistics in medicine}, 24\penalty0 (24), 2005.

\bibitem[Burger et~al.(2025)Burger, Chopard, Londschien, Sergeev, Y{\`e}che, Kuznetsova, Faltys, Gerdes, Leshetkina, B{\"u}hlmann, et~al.]{burger2025foundation}
Manuel Burger, Daphn{\'e} Chopard, Malte Londschien, Fedor Sergeev, Hugo Y{\`e}che, Rita Kuznetsova, Martin Faltys, Eike Gerdes, Polina Leshetkina, Peter B{\"u}hlmann, et~al.
\newblock A foundation model for intensive care: Unlocking generalization across tasks and domains at scale.
\newblock \emph{medRxiv}, pages 2025--07, 2025.

\bibitem[Burkhart et~al.(2025)Burkhart, Ramadan, Liao, Chhikara, Rojas, Parker, and Beaulieu-Jones]{burkhart2025foundation}
Michael~C Burkhart, Bashar Ramadan, Zewei Liao, Kaveri Chhikara, Juan~C Rojas, William~F Parker, and Brett~K Beaulieu-Jones.
\newblock Foundation models for electronic health records: representation dynamics and transferability.
\newblock \emph{arXiv preprint arXiv:2504.10422}, 2025.

\bibitem[Chen et~al.(2024)]{chen2024introduction}
George~H Chen et~al.
\newblock An introduction to deep survival analysis models for predicting time-to-event outcomes.
\newblock \emph{Foundations and Trends{\textregistered} in Machine Learning}, 17\penalty0 (6):\penalty0 921--1100, 2024.

\bibitem[Chen and Guestrin(2016)]{chen2016xgboost}
Tianqi Chen and Carlos Guestrin.
\newblock Xgboost: A scalable tree boosting system.
\newblock In \emph{Proceedings of the 22nd ACM SIGKDD International Conference on Knowledge Discovery and Data Mining}, KDD ’16, page 785–794. ACM, August 2016.

\bibitem[Du et~al.(2016)Du, Dai, Trivedi, Upadhyay, Gomez-Rodriguez, and Song]{du2016recurrent}
Nan Du, Hanjun Dai, Rakshit Trivedi, Utkarsh Upadhyay, Manuel Gomez-Rodriguez, and Le~Song.
\newblock Recurrent marked temporal point processes: Embedding event history to vector.
\newblock In \emph{Proceedings of the 22nd ACM SIGKDD international conference on knowledge discovery and data mining}, pages 1555--1564, 2016.

\bibitem[Fallahpour et~al.(2024)Fallahpour, Alinoori, Ye, Cao, Afkanpour, and Krishnan]{fallahpour2024ehrmamba}
Adibvafa Fallahpour, Mahshid Alinoori, Wenqian Ye, Xu~Cao, Arash Afkanpour, and Amrit Krishnan.
\newblock Ehrmamba: Towards generalizable and scalable foundation models for electronic health records.
\newblock \emph{arXiv preprint arXiv:2405.14567}, 2024.

\bibitem[Gadd et~al.(2025)Gadd, Gokhale, Acharya, Cooper, Crowe, Fitzsimmons, Jackson, Nirantharakumar, Yau, and collaborative]{gadd2025survivehr}
Charles Gadd, Krishna Gokhale, Aditya Acharya, Jennifer Cooper, Francesca Crowe, Leah Fitzsimmons, Thomas Jackson, Krishnarajah Nirantharakumar, Christopher Yau, and OPTIMAL collaborative.
\newblock Survivehr: a competing risks, time-to-event foundation model for multiple long-term conditions from primary care electronic health records.
\newblock \emph{medRxiv}, pages 2025--08, 2025.

\bibitem[Gu and Dao(2024)]{gu2024mamba}
Albert Gu and Tri Dao.
\newblock Mamba: Linear-time sequence modeling with selective state spaces.
\newblock In \emph{First conference on language modeling}, 2024.

\bibitem[Guo et~al.(2023)Guo, Steinberg, Fleming, Posada, Lemmon, Pfohl, Shah, Fries, and Sung]{guo2023ehr}
Lin~Lawrence Guo, Ethan Steinberg, Scott~Lanyon Fleming, Jose Posada, Joshua Lemmon, Stephen~R Pfohl, Nigam Shah, Jason Fries, and Lillian Sung.
\newblock Ehr foundation models improve robustness in the presence of temporal distribution shift.
\newblock \emph{Scientific Reports}, 13\penalty0 (1):\penalty0 3767, 2023.

\bibitem[Guo et~al.(2024)Guo, Fries, Steinberg, Fleming, Morse, Aftandilian, Posada, Shah, and Sung]{guo2024multi}
Lin~Lawrence Guo, Jason Fries, Ethan Steinberg, Scott~Lanyon Fleming, Keith Morse, Catherine Aftandilian, Jose Posada, Nigam Shah, and Lillian Sung.
\newblock A multi-center study on the adaptability of a shared foundation model for electronic health records.
\newblock \emph{NPJ digital medicine}, 7\penalty0 (1):\penalty0 171, 2024.

\bibitem[Guo et~al.(2026)Guo, Arciniegas, Lee, Yan, Tomlinson, Fries, and Sung]{guo2026tokenization}
Lin~Lawrence Guo, Santiago~Eduardo Arciniegas, Joseph~Jihyung Lee, Adam~Paul Yan, George Tomlinson, Jason Fries, and Lillian Sung.
\newblock Tokenization tradeoffs in structured ehr foundation models.
\newblock \emph{arXiv preprint arXiv:2603.15644}, 2026.

\bibitem[Hegselmann et~al.(2025)Hegselmann, von Arnim, Rheude, Kronenberg, Sontag, Hindricks, Eils, and Wild]{hegselmann2025large}
Stefan Hegselmann, Georg von Arnim, Tillmann Rheude, Noel Kronenberg, David Sontag, Gerhard Hindricks, Roland Eils, and Benjamin Wild.
\newblock Large language models are powerful electronic health record encoders.
\newblock \emph{arXiv preprint arXiv:2502.17403}, 2025.

\bibitem[Hur et~al.(2023)Hur, Oh, Kim, Kim, Lee, Cho, Moon, Kim, Atallah, and Choi]{hur2023genhpf}
Kyunghoon Hur, Jungwoo Oh, Junu Kim, Jiyoun Kim, Min~Jae Lee, Eunbyeol Cho, Seong-Eun Moon, Young-Hak Kim, Louis Atallah, and Edward Choi.
\newblock Genhpf: General healthcare predictive framework for multi-task multi-source learning.
\newblock \emph{IEEE Journal of Biomedical and Health Informatics}, 28\penalty0 (1):\penalty0 502--513, 2023.

\bibitem[Im et~al.(2025)Im, Oh, and Choi]{im2025labtop}
Sujeong Im, Jungwoo Oh, and Edward Choi.
\newblock Labtop: A unified model for lab test outcome prediction on electronic health records.
\newblock \emph{arXiv preprint arXiv:2502.14259}, 2025.

\bibitem[Jeanselme et~al.(2023)Jeanselme, Yoon, Tom, and Barrett]{jeanselme2023neural}
Vincent Jeanselme, Chang~Ho Yoon, Brian Tom, and Jessica Barrett.
\newblock Neural fine-gray: Monotonic neural networks for competing risks.
\newblock In \emph{Conference on Health, Inference, and Learning}, pages 379--392. PMLR, 2023.

\bibitem[Johnson et~al.(2023)Johnson, Bulgarelli, Shen, Gayles, Shammout, Horng, Pollard, Hao, Moody, Gow, et~al.]{johnson2023mimic}
Alistair~EW Johnson, Lucas Bulgarelli, Lu~Shen, Alvin Gayles, Ayad Shammout, Steven Horng, Tom~J Pollard, Sicheng Hao, Benjamin Moody, Brian Gow, et~al.
\newblock Mimic-iv, a freely accessible electronic health record dataset.
\newblock \emph{Scientific data}, 10\penalty0 (1):\penalty0 1, 2023.

\bibitem[Lee et~al.(2018)Lee, Zame, Yoon, and Van Der~Schaar]{lee2018deephit}
Changhee Lee, William Zame, Jinsung Yoon, and Mihaela Van Der~Schaar.
\newblock Deephit: A deep learning approach to survival analysis with competing risks.
\newblock In \emph{Proceedings of the AAAI conference on artificial intelligence}, volume~32, 2018.

\bibitem[Lee et~al.(2025)Lee, Jain, Chen, Ono, Biswas, Rudas, Fang, and Chiang]{lee2025clinical}
Simon~A Lee, Sujay Jain, Alex Chen, Kyoka Ono, Arabdha Biswas, {\'A}kos Rudas, Jennifer Fang, and Jeffrey~N Chiang.
\newblock Clinical decision support using pseudo-notes from multiple streams of ehr data.
\newblock \emph{npj Digital Medicine}, 8\penalty0 (1):\penalty0 394, 2025.

\bibitem[Li et~al.(2020)Li, Rao, Solares, Hassaine, Ramakrishnan, Canoy, Zhu, Rahimi, and Salimi-Khorshidi]{li2020behrt}
Yikuan Li, Shishir Rao, Jos{\'e} Roberto~Ayala Solares, Abdelaali Hassaine, Rema Ramakrishnan, Dexter Canoy, Yajie Zhu, Kazem Rahimi, and Gholamreza Salimi-Khorshidi.
\newblock Behrt: transformer for electronic health records.
\newblock \emph{Scientific reports}, 10\penalty0 (1):\penalty0 7155, 2020.

\bibitem[Mei and Eisner(2017)]{mei2017neural}
Hongyuan Mei and Jason~M Eisner.
\newblock The neural hawkes process: A neurally self-modulating multivariate point process.
\newblock \emph{Advances in neural information processing systems}, 30, 2017.

\bibitem[Odgaard et~al.(2024)Odgaard, Klein, Thysen, Jimenez-Solem, Sillesen, and Nielsen]{odgaard2024core}
Mikkel Odgaard, Kiril~Vadimovic Klein, Sanne~M{\o}ller Thysen, Espen Jimenez-Solem, Martin Sillesen, and Mads Nielsen.
\newblock Core-behrt: A carefully optimized and rigorously evaluated behrt.
\newblock \emph{arXiv preprint arXiv:2404.15201}, 2024.

\bibitem[Omi et~al.(2019)Omi, Aihara, et~al.]{omi2019fully}
Takahiro Omi, Kazuyuki Aihara, et~al.
\newblock Fully neural network based model for general temporal point processes.
\newblock \emph{Advances in neural information processing systems}, 32, 2019.

\bibitem[Pang et~al.(2021{\natexlab{a}})Pang, Jiang, Kalluri, Spotnitz, Chen, Perotte, and Natarajan]{pang2021cehr}
Chao Pang, Xinzhuo Jiang, Krishna~S Kalluri, Matthew Spotnitz, RuiJun Chen, Adler Perotte, and Karthik Natarajan.
\newblock Cehr-bert: Incorporating temporal information from structured ehr data to improve prediction tasks.
\newblock In \emph{Machine learning for health}, pages 239--260. PMLR, 2021{\natexlab{a}}.

\bibitem[Pang et~al.(2021{\natexlab{b}})Pang, Jiang, Kalluri, Spotnitz, Chen, Perotte, and Natarajan]{pmlr-v158-pang21a}
Chao Pang, Xinzhuo Jiang, Krishna~S. Kalluri, Matthew Spotnitz, RuiJun Chen, Adler Perotte, and Karthik Natarajan.
\newblock Cehr-bert: Incorporating temporal information from structured ehr data to improve prediction tasks.
\newblock In Subhrajit Roy, Stephen Pfohl, Emma Rocheteau, Girmaw~Abebe Tadesse, Luis Oala, Fabian Falck, Yuyin Zhou, Liyue Shen, Ghada Zamzmi, Purity Mugambi, Ayah Zirikly, Matthew B.~A. McDermott, and Emily Alsentzer, editors, \emph{Proceedings of Machine Learning for Health}, volume 158 of \emph{Proceedings of Machine Learning Research}, pages 239--260. PMLR, 04 Dec 2021{\natexlab{b}}.
\newblock URL \url{https://proceedings.mlr.press/v158/pang21a.html}.

\bibitem[Pang et~al.(2024)Pang, Jiang, Pavinkurve, Kalluri, Minto, Patterson, Zhang, Hripcsak, G{\"u}rsoy, Elhadad, et~al.]{pang2024cehr}
Chao Pang, Xinzhuo Jiang, Nishanth~Parameshwar Pavinkurve, Krishna~S Kalluri, Elise~L Minto, Jason Patterson, Linying Zhang, George Hripcsak, Gamze G{\"u}rsoy, No{\'e}mie Elhadad, et~al.
\newblock Cehr-gpt: Generating electronic health records with chronological patient timelines.
\newblock \emph{arXiv preprint arXiv:2402.04400}, 2024.

\bibitem[Pang et~al.(2025)Pang, Jeanselme, Choi, Jiang, Jing, Kashyap, Kobayashi, Li, Pollet, Natarajan, et~al.]{pang2025fomoh}
Chao Pang, Vincent Jeanselme, Young~Sang Choi, Xinzhuo Jiang, Zilin Jing, Aparajita Kashyap, Yuta Kobayashi, Yanwei Li, Florent Pollet, Karthik Natarajan, et~al.
\newblock Fomoh: A clinically meaningful foundation model evaluation for structured electronic health records.
\newblock \emph{arXiv preprint arXiv:2505.16941}, 2025.

\bibitem[Renc et~al.(2024)Renc, Jia, Samir, Was, Li, Bates, and Sitek]{renc2024zero}
Pawel Renc, Yugang Jia, Anthony~E Samir, Jaroslaw Was, Quanzheng Li, David~W Bates, and Arkadiusz Sitek.
\newblock Zero shot health trajectory prediction using transformer.
\newblock \emph{NPJ Digital Medicine}, 7\penalty0 (1):\penalty0 256, 2024.

\bibitem[Shmatko et~al.(2025)Shmatko, Jung, Gaurav, Brunak, Mortensen, Birney, Fitzgerald, and Gerstung]{shmatko2025learning}
Artem Shmatko, Alexander~Wolfgang Jung, Kumar Gaurav, S{\o}ren Brunak, Laust~Hvas Mortensen, Ewan Birney, Tom Fitzgerald, and Moritz Gerstung.
\newblock Learning the natural history of human disease with generative transformers.
\newblock \emph{Nature}, 647\penalty0 (8088):\penalty0 248--256, 2025.

\bibitem[Steinberg et~al.(2024)Steinberg, Xu, Fries, and Shah]{steinberg2024motor}
Ethan Steinberg, Yizhe Xu, Jason~Alan Fries, and Nigam Shah.
\newblock {MOTOR}: A time-to-event foundation model for structured medical records.
\newblock In \emph{The Twelfth International Conference on Learning Representations}, 2024.
\newblock URL \url{https://openreview.net/forum?id=NialiwI2V6}.

\bibitem[Vaswani et~al.(2017)Vaswani, Shazeer, Parmar, Uszkoreit, Jones, Gomez, Kaiser, and Polosukhin]{vaswani2017attention}
Ashish Vaswani, Noam Shazeer, Niki Parmar, Jakob Uszkoreit, Llion Jones, Aidan~N Gomez, {\L}ukasz Kaiser, and Illia Polosukhin.
\newblock Attention is all you need.
\newblock \emph{Advances in neural information processing systems}, 30, 2017.

\bibitem[Wornow et~al.(2024)Wornow, Bedi, Hernandez, Steinberg, Fries, R{\'e}, Koyejo, and Shah]{wornow2024context}
Michael Wornow, Suhana Bedi, Miguel Angel~Fuentes Hernandez, Ethan Steinberg, Jason~Alan Fries, Christopher R{\'e}, Sanmi Koyejo, and Nigam~H Shah.
\newblock Context clues: Evaluating long context models for clinical prediction tasks on ehrs.
\newblock \emph{arXiv preprint arXiv:2412.16178}, 2024.

\end{thebibliography}

%%%%%%%%%%%%%%%%%%%%%%%%%%%%%%%%%%%%%%%%%%%%%%%%%%%%%%%%%%%%%%%%%%%%%%%%%%%%%%%
%%%%%%%%%%%%%%%%%%%%%%%%%%%%%%%%%%%%%%%%%%%%%%%%%%%%%%%%%%%%%%%%%%%%%%%%%%%%%%%
% APPENDIX
%%%%%%%%%%%%%%%%%%%%%%%%%%%%%%%%%%%%%%%%%%%%%%%%%%%%%%%%%%%%%%%%%%%%%%%%%%%%%%%
%%%%%%%%%%%%%%%%%%%%%%%%%%%%%%%%%%%%%%%%%%%%%%%%%%%%%%%%%%%%%%%%%%%%%%%%%%%%%%%
\newpage
\appendix

\section{Comparison of Different Pretraining Losses}

\label{fig:comparison}
\begin{table}[ht]
\centering
\caption{\textbf{Comparison of pretraining objectives of existing EHR foundation models} "Mixed Event Type" denotes if the model includes events with and without associated numerical values in the loss function. "Time–Value Dependence" means that the model considers the dependence between time and the value of the same event. "Concurrent Events" denotes the ability to predict co-occurring events.}
\label{tab:perf}
\small
\renewcommand{\arraystretch}{1.3}
\begin{tabular}{l@{\hskip 0.3cm}cc@{\hskip 0.3cm}cc}
\toprule
\textbf{Model} & Mixed Event Types & Time–Value Dependence & Concurrent Events  \\
\midrule
Context Clue \citep{wornow2024context} & \gcheck & \redx & \gcheck \\
MOTOR \citep{steinberg2024motor} & \redx & \redx & \gcheck \\
SurvivalEHR \citep{gadd2025survivehr} & \gcheck  & \redx & \redx \\
ICareFM \citep{burger2025foundation} & \redx& \gcheck & \gcheck \\
\midrule
\rowcolor{gray!15}
\textbf{ORA (ours)} & \gcheck & \gcheck & \gcheck \\
\bottomrule
\end{tabular}
\end{table}
\FloatBarrier

% \begin{table}[t]
% \centering
% \caption{\textbf{Comparison of pretraining objectives of existing EHR foundation models} "Mixed Event Type" denotes if the model includes events with and without associated numerical values in the loss function. "Time–Value Dependence" means that the model considers the dependence between time and the value of the same event. "Concurrent Events" denotes the ability to predict co-occurring events.}
% \label{tab:perf}
% \small
% \renewcommand{\arraystretch}{1.3}

% \begin{tabular}{l@{\hskip 0.3cm}cc@{\hskip 0.3cm}cc}
% \toprule
% & \multicolumn{2}{c}{\textbf{Event Type}} & 
% \textbf{Time–Value Dependence} & \textbf{Time–Value Dependence} & \textbf{Time–Value Dependence} & \textbf{Concurrent Events} \\
% \cmidrule(lr){2-3} \cmidrule(lr){4-4} \cmidrule(lr){5-5}

% \textbf{Model} & Mixed Event Types & Time–Value Dependence & Concurrent Events  \\
% \midrule
% Context Clue \citep{wornow2024context} & \gcheck & \redx & \gcheck \\
% MOTOR \citep{steinberg2024motor} & \redx & \redx & \gcheck \\
% SurvivalEHR \citep{gadd2025survivehr} & \gcheck  & \redx & \redx \\
% ICareFM \citep{burger2025foundation} & \redx& \gcheck & \gcheck \\
% \midrule
% \rowcolor{gray!15}
% \textbf{ORA (ours)} & \gcheck & \gcheck & \gcheck \\
% \bottomrule
% \end{tabular}
% \end{table}

\section{Experimental Setting}

\subsection{Datasets}
\label{app:datasets}

% Preamble:
% \usepackage{booktabs}
% \usepackage{siunitx}
% \sisetup{group-separator={,}, group-minimum-digits=4}

\begin{table}[ht]
\centering
\caption{Dataset Split of MIMIC-IV and \cumc}
\label{tab:dataset_split_details}
    \begin{tabular}{ccccc}
        \toprule
        \textbf{Split Name} & \multicolumn{2}{c}{\textbf{MIMIC-IV}} & \multicolumn{2}{c}{\textbf{\cumc}} \\
        \cmidrule(lr){2-3} \cmidrule(lr){4-5}
        & {\# Patients} & {\# Events} & {\# Patients} & {\# Events} \\
        \midrule
        Training Set   & 291,702 & 579,667,440  & 3,996,578 & 1,562,316,866\\
        Tuning Set & 36463  &  71,789,378 & 705,279 & 273,247,565 \\
        Test Set       & 36462  & 71,290,747 & 2,015,082 & 781,462,164 \\
        \bottomrule
    \end{tabular}
\end{table}
\FloatBarrier

\subsection{Cohort Definition}
\subsubsection{Outcome definition}
We construct three outcome tasks using the same definitions as in \citep{pang2025fomoh}. The detailed inclusion criteria can be found in section 3.2 of the referenced paper.

\subsubsection{Phenotype Definition}
\label{app:classification}
For each disease, we define a set of at-risk events as cohort inclusion criteria and a set of case events to determine patients' labels. Compared with using a set of ICD codes, this method adds more task difficulty and is more clinically meaningful. 

\begin{figure}[!ht]
\centering
\includegraphics[width=0.7\textwidth]{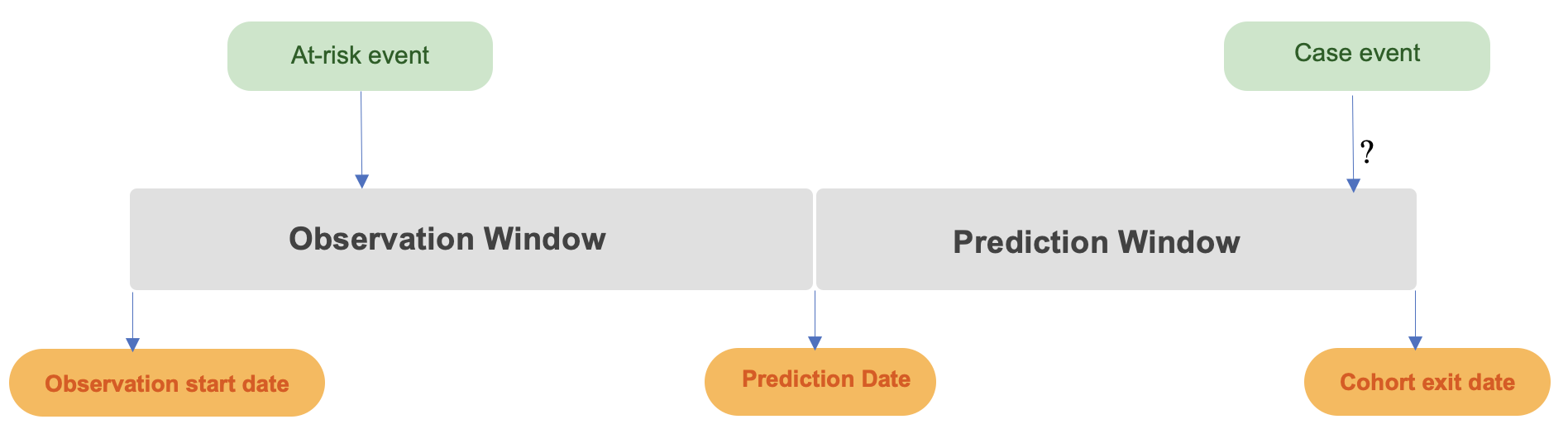}
\caption{Visualization of cohort definition. The detailed definition of at-risk, prediction time, and case events can be found in Appendix B of \citet{pang2025fomoh}.}
\end{figure}

\subsubsection{Regression Definition}
\label{app:regression}
We define regression tasks as predicting the lab test after 4 hours of the prediction time. Each lab  event can be identified with a corresponding code in the following table:

\begin{table}[!ht]
  \centering
  \caption{Corresponding medical codes for all regression tasks on MIMIC and \cumc }
  
  \begin{tabular}{lccc}
    \toprule
    \textbf{Task} & \textbf{MIMIC-IV} & \textbf{\cumc(OMOP)} \\
    \midrule

    % \textbf{Bilirubin} & 50885  & LOINC/1968-7 &  \\
    \textbf{Creatinine} &  50912 & LOINC/2160-0 &  \\
    \textbf{Platelets}  & 51265  & LOINC/26515-7  &  \\
    \textbf{Oxygen} & 50821 ($PaO_2$) & LOINC/2708-6 ($SpO_2$) & \\
     \textbf{Glucose} & 50931  & LOINC/2339-0 & \\ 

    \bottomrule
  \end{tabular}
\end{table}

% We only include non-missing value in our predictions

\subsection{Model}\label{app:model}

\subsubsection{Tokenizer}
\label{app:tokenizer}
We use an entropy-based filter similar to \citep{steinberg2024motor}. Instead of taking the most frequent events in the vocabulary, we select codes with the highest entropy over the whole dataset. Define $p(m)$ as the probability that the code appears in each patient. The entropy calculation is as follows:

\begin{align*}
    H(m) = -p(m){\rm log}p(m)
\end{align*}

If the dataset has an ontology mapping, we can calculate the conditional entropy of any code $m$ relative to its parent $n$. Suppose $p(m,n^+)$ denotes the probability that both $m$ and $n$ appear per patient and $p(m,n^-)$ be the probability that only $m$ appears in the patient. $p(m)=p(m,n^+)+p(m,n^-)$, the conditional entropy is as follows:

\begin{align*}
    H(m|n) = -p(m,n^-) {\rm log} \frac{p(m,n^-)}{p(m)} -p(m,n^+) {\rm log} \frac{p(m,n^+)}{p(m)}
\end{align*}

\subsubsection{Model Configurations}
\label{sec:model_configurations}
For a fair comparison across different models, we set the parameter size for all models to around 120M. Within each architecture, we also use the same configuration files for different losses, which is specified as follows:
\label{app:hyperparam}
\begin{table}[t]
\centering
\caption{Model configurations.}
\label{tab:model_configs}
\begin{tabular}{l l c}
\toprule
\textbf{Model} & \textbf{Configuration} & \textbf{Value} \\
\midrule

Transformer   & context length     & $8192$ \\
      & learning rate   & $1\mathrm{e}{-5}$ \\
      & dim model       & $768$ \\
      & intermedicate size & $3072$ \\
      & num layers      & $11$ \\
      & num heads       & $12$ \\
      & \textbf{Total Parameters} & \textbf{119M} \\
\midrule

Mamba & context length     & $8192$ \\
      & learning rate   & $2\mathrm{e}{-4}$ \\
      & dim model       & $768$ \\
      & intermedicate size & $1536$ \\
      & num layers      & $28$ \\
      & num heads      & $16$ \\
      & \textbf{Total Parameters} & \textbf{120M} \\
\bottomrule
\end{tabular}
\end{table}

\newpage
\subsubsection{Efficient Prediction Head}
\label{app:compression}

The last layer of Transformer and Mamba produces an embedding matrix $X \in R^{\times D}$. We need to project it into the final probability matrix $P(x)$ for both nonnumerical codes (eg, diagnosis) and numerical codes with continuous values (eg, lab test). As mentioned in the main paper, we adopt a two-stage projection to be parameter efficient. $D$ indicates the embedding dimension after Transformer or Mamba. $D_2$ is the hidden dimension after the first projection layer. $M_{non}$ and $M_{num}$ represent nonnumerical codes and numerical codes for pretraining, respectively. The visualization is as follows:

\begin{figure}[h]
    \centering
    \includegraphics[width=0.9\linewidth]{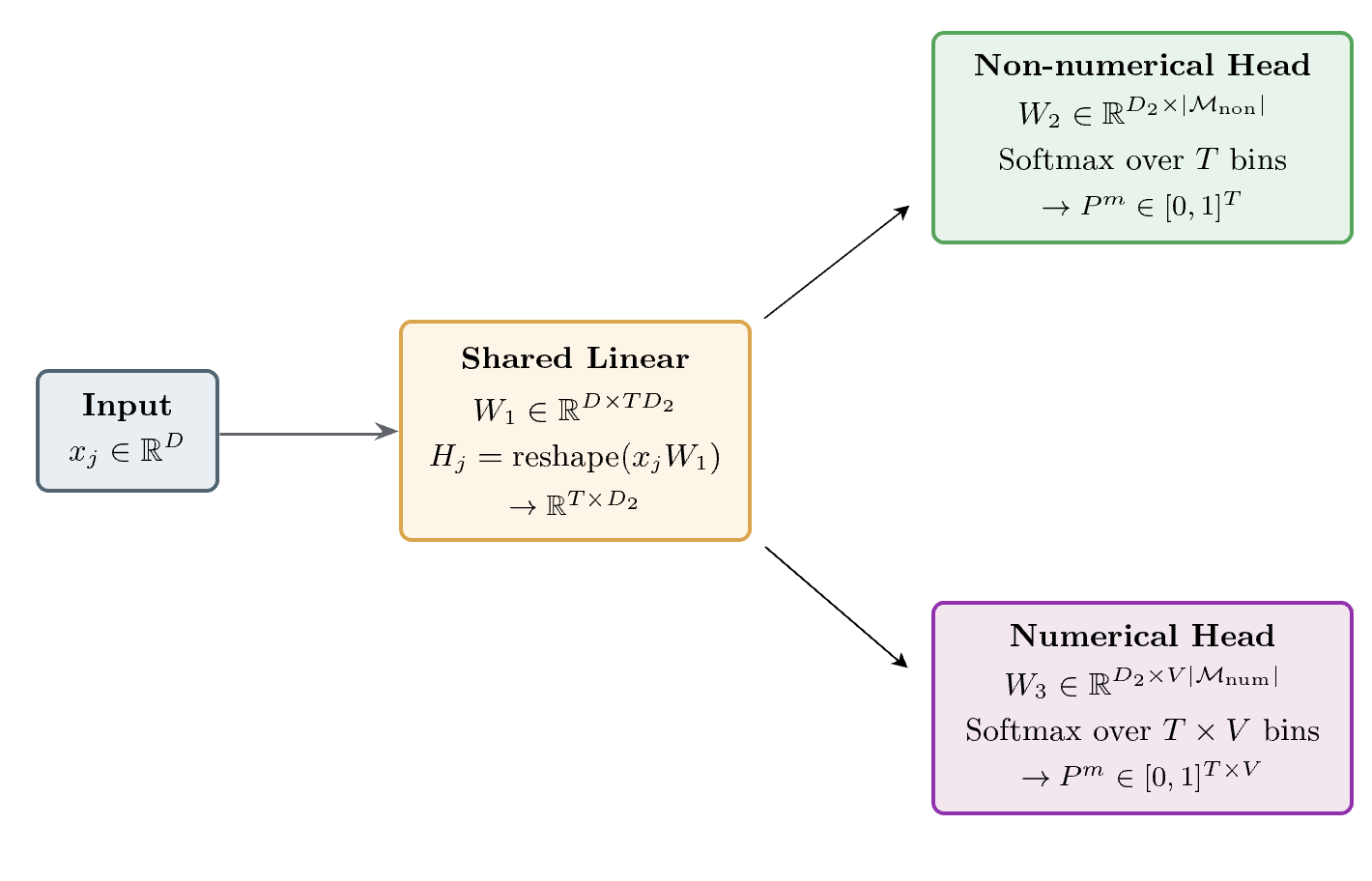} % TO 
    \caption{Efficient prediction head: We share the same projection layer for both nonnumerical and numerical codes and then use two different projection heads to output the separate probability matrices.}
    \label{fig:projection}
\end{figure}

\FloatBarrier

\subsection{Hyperparameter Search for Discretized Intervals}
\label{sec:hyperparameter}
Following prior choices for time and value discretization~\citep{renc2024zero,steinberg2024motor}, we initially set $T=8$ and $V=10$. We further conducted ablations over $T \in \{8,16\}$ and $V \in \{5,10\}$ by pretraining four models on each dataset and evaluating them across all downstream tasks. For each task, we ranked the four settings by performance (from best to worst) and reported the average rank across tasks for each setting. We report the ablation study results on MIMIC-IV in Table~\ref{tab:tv_ablation}. $T=8, V=5$ is the strongest and most stable setting, achieving the lowest average rank. Although coarser discretization uses fewer bins and performs better in downstream evaluation, it may be less precise for capturing fine-grained values during zero-shot generation. In contrast, using more bins increases the number of parameters in the final projection head and may require longer training for stable convergence.

\begin{table}[ht]
\centering
\caption{Ablation study of time and value discretization on MIMIC-IV. We report the average rank across all downstream tasks. Lower rank reflects better performance.}
\label{tab:tv_ablation}
\begin{tabular}{lcc}
\toprule
\textbf{Setting} & \textbf{Average rank} \\
\midrule
$T=8,\ V=5$   & $1.33$ \\
$T=16,\ V=10$ & $2.58$ \\
$T=16,\ V=5$   & $2.75$ \\
$T=8,\ V=10$  & $3.33$ \\
\bottomrule
\end{tabular}
\end{table}

\FloatBarrier

\section{Averaged Performance for Mamba Backbones}
\label{sec:mamba_results}
To test whether our loss function is robust across different backbones, we further pre-trained all models with the same Mamba backbone and evaluated them on the MIMIC-IV datasets. We present the full comparison of all models in Figure~\ref{fig:mimic_baseline_mamba} Similarly, we also fix the tokenizer and the backbone to compare how different loss functions impact the downstream task. The results are reported in Figure~\ref{fig:mimic_ablation_mamba}.

\begin{figure}[ht]
    \centering
    \includegraphics[width=\linewidth]{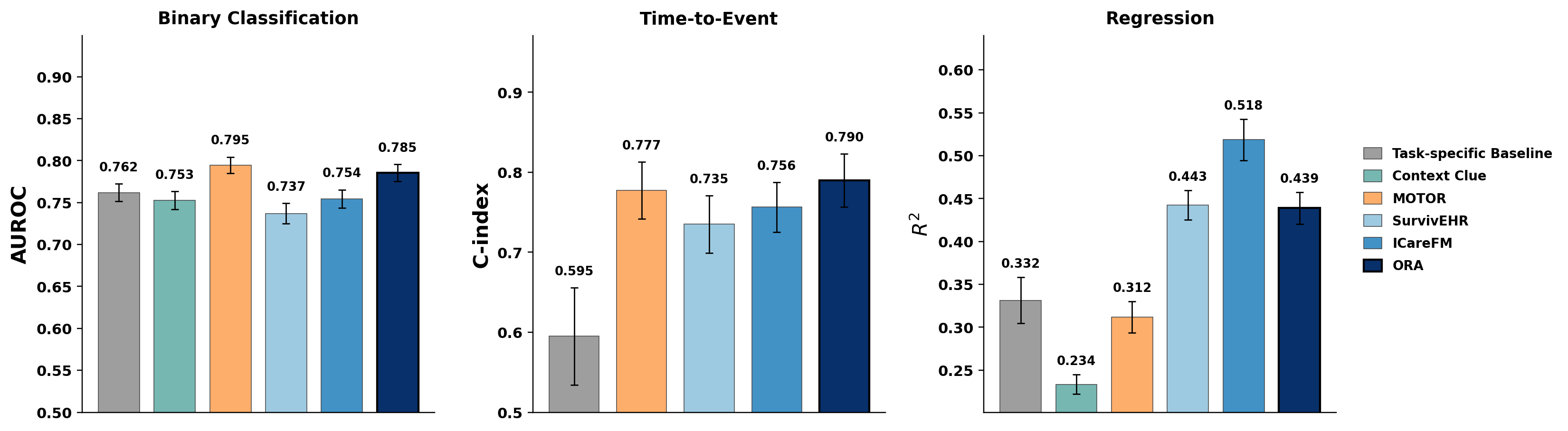}
\caption{\textbf{ORA Improves Generalization Across Classification, Time-to-Event, and Regression Tasks.} Average performance on MIMIC-IV shows that jointly modeling event timing and values yields consistently stronger representations. ORA outperforms different foundation models across both binary classification and TTE tasks and remains competitive in regression results. All models used the \textbf{Mamba} backbone.
% on MIMIC-IV and \cumc~ datasets. All models used the \textbf{Transformer} architecture.
}
    \label{fig:mimic_baseline_mamba}
\end{figure}

\begin{figure}[ht]
    \centering
    \includegraphics[width=\linewidth]{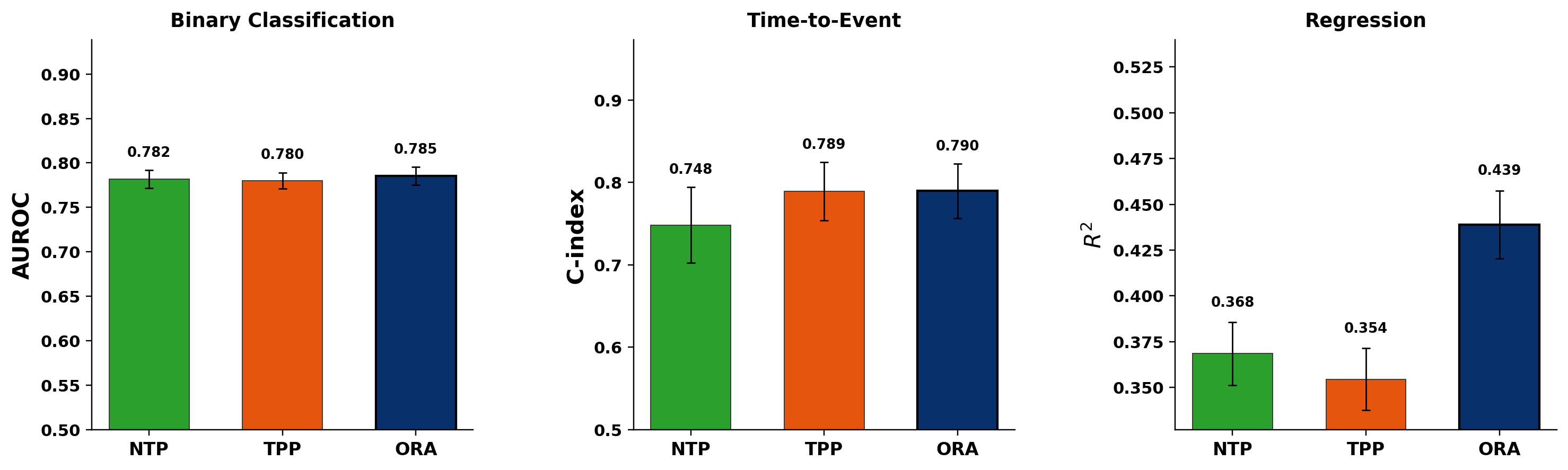}
\caption{\textbf{Pretraining Objective Drives Downstream Performance Across Task Families on MIMIC-IV using the Mamba Backbone.}
Under a fixed tokenizer and Mamba backbone, jointly modeling time-to-event and values yields consistent gains over NTP and TPP objectives, demonstrating that improvements arise from the loss formulation rather than model capacity or input representation.}
    \label{fig:mimic_ablation_mamba}
\end{figure}

\clearpage
\section{Full Experiment results}
\label{app:results}
This section details the performance for each backbone, loss, and task across both datasets.

\subsection{MIMIC-IV}
We evaluate all models on 6 classification tasks, 3 time-to-event tasks, and 4 regression tasks on MIMIC-IV. Note that we don't extract the Celiac cohort in MIMIC-IV.
\subsubsection{Classification}

\begin{table}[!ht]
  \centering
  \caption{AUROC of binary classification tasks for MIMIC-IV patients. \textit{Larger is better.}}
  
  \begin{tabular}{lcccccc}
    \toprule
    \textbf{Model} & \textbf{Readmission} & \textbf{LOS} & \textbf{Mortality} & \textbf{AMI} & \textbf{MASLD} & \textbf{Stroke} \\
    \midrule

    \textbf{XGBoost}            & \makecell{0.726\\(0.003)} & \makecell{0.748\\(0.003)} & \makecell{0.882\\(0.005)} & \makecell{0.807\\(0.008)} & \makecell{0.700\\(0.020)} & \makecell{0.708\\(0.023)} \\

    \textbf{CC-Llama}           & \makecell{0.730\\(0.003)} & \makecell{0.781\\(0.003)} & \makecell{0.901\\(0.005)} &
    \makecell{0.796\\(0.010)} & \makecell{0.663\\(0.020)} & \makecell{0.696\\(0.025)} \\

    % updated AMI / MASLD / Stroke from your new "mamba" row
    \textbf{CC-Mamba}           & \makecell{0.731\\(0.003)} & \makecell{0.779\\(0.003)} & \makecell{0.896\\(0.006)} &
    \makecell{0.783\\(0.009)} & \makecell{0.661\\(0.019)} & \makecell{0.665\\(0.024)} \\

    \textbf{MOTOR}              & \makecell{0.743\\(0.002)} & \makecell{0.833\\(0.002)} & \makecell{0.954\\(0.002)} & \makecell{0.825\\(0.008)} & \makecell{0.703\\(0.021)} & \makecell{0.710\\(0.019)} \\

    \midrule\midrule % line between MOTOR and Transformer-NTP

    \textbf{Transformer-NTP}    & \makecell{0.728\\(0.003)} & \makecell{0.808\\(0.002)} & \makecell{0.926\\(0.004)} & \makecell{0.810\\(0.008)} & \makecell{0.691\\(0.020)} & \makecell{0.691\\(0.025)} \\
    \textbf{Transformer-TPP} & \makecell{0.744\\(0.004)} & \makecell{0.839\\(0.002)} & \makecell{0.963\\(0.003)} & \makecell{0.818\\(0.007)} & \makecell{0.705\\(0.018)} & \makecell{0.700\\(0.020)} \\
    % \textbf{Transformer-Numerical} & \makecell{0.736\\(0.003)}  & \makecell{0.802\\(0.002)} & \makecell{0.955\\(0.003)} &  \makecell{0.816\\(0.006)} & \makecell{0.697\\(0.016)} & \makecell{0.686\\(0.021)} \\
    \textbf{Transformer-SurvivEHR}              & \makecell{0.714\\(0.003)} & \makecell{0.734\\(0.003)} & \makecell{0.898\\(0.005)} & \makecell{0.803\\(0.007)} & \makecell{0.682\\(0.016)} & \makecell{0.603\\(0.022)} \\

    \textbf{Transformer-ICareFM}              & \makecell{0.719\\(0.004)} & \makecell{0.796\\(0.003)} & \makecell{0.935\\(0.005)} & \makecell{0.802\\(0.007)} & \makecell{0.659\\(0.021)} & \makecell{0.652\\(0.023)} \\
    \textbf{Transformer-ORA}   & \makecell{0.745\\(0.003)} & \textbf{\makecell{0.841\\(0.002)}} & \makecell{\textbf{0.965}\\\textbf{(0.002)}} & \makecell{0.827\\(0.007)} & \makecell{0.714\\(0.017)} & \makecell{0.711\\(0.018)} \\
% \textbf{ORA vs NTP} & \makecell{+2.34\%} & \makecell{+4.08\%} & \makecell{+4.21\%} & \makecell{+2.10\%} & \makecell{+3.33\%} & \makecell{+2.89\%} \\

    \midrule\midrule % line between Transformer-ORA and Mamba-NTP

    \textbf{Mamba-NTP}          & \makecell{0.732\\(0.004)} & \makecell{0.813\\(0.002)} & \makecell{0.933\\(0.004)} & \makecell{0.814\\(0.008)} & \makecell{0.679\\(0.020)} & \makecell{\textbf{0.719}\\\textbf{(0.022)}} \\
    \textbf{Mamba-TPP}       & \makecell{0.741\\(0.003)} & \makecell{0.809\\(0.003)} & \makecell{0.934\\(0.004)} & \makecell{0.822\\(0.007)} & \makecell{0.709\\(0.020)} & \makecell{0.666\\(0.017)} \\
    \textbf{Mamba-SurvivEHR}    & \makecell{0.714\\(0.004)} & \makecell{0.743\\(0.003)} & \makecell{0.900\\(0.006)} & \makecell{0.786\\(0.007)} & \makecell{0.643\\(0.021)} & \makecell{0.635\\(0.029)} \\

    \textbf{Mamba-ICareFM}      & \makecell{0.728\\(0.003)} & \makecell{0.787\\(0.003)} & \makecell{0.931\\(0.004)} & \makecell{0.792\\(0.007)} & \makecell{0.661\\(0.021)} & \makecell{0.627\\(0.024)} \\
    \textbf{Mamba-ORA}         & \textbf{\makecell{0.747\\(0.003)}} & \makecell{0.812\\(0.002)} & \makecell{0.939\\(0.004)} & \makecell{\textbf{0.832}\\\textbf{(0.007)}} & \makecell{\textbf{0.722}\\\textbf{(0.020)}} & \makecell{0.660\\(0.022)} \\
% \textbf{ORA vs NTP} & \makecell{+2.05\%} & \makecell{-0.12\%} & \makecell{+0.64\%} & \makecell{+2.21\%} & \makecell{+6.33\%} & \makecell{-7.37\%} \\

    \bottomrule
  \end{tabular}
\end{table}

\newpage
\subsubsection{Time-to-event}
\begin{table}[!ht]
  \centering
  \caption{C-index of time-to-event prediction tasks for MIMIC-IV patients. \textit{Larger is better.}}

  \begin{tabular}{lccc}
    \toprule
    \textbf{Model} & \textbf{AMI} & \textbf{MASLD} & \textbf{Stroke} \\
    \midrule

    \textbf{Deephit}
    & \makecell{0.651\\(0.029)}
    & \makecell{0.560\\(0.024)}
    & \makecell{0.574\\(0.085)} \\

    \textbf{CC-Llama}
    & \makecell{0.810\\(0.029)}
    & \makecell{0.688\\(0.022)}
    & \makecell{0.843\\(0.038)} \\

    \textbf{MOTOR}
    & \makecell{0.811\\(0.030)}
    & \makecell{0.689\\(0.020)}
    & \makecell{0.832\\(0.041)} \\

    \midrule\midrule

    \textbf{Transformer-NTP}
    & \makecell{0.796\\(0.030)}
    & \makecell{0.706\\(0.018)}
    & \makecell{0.702\\(0.056)} \\

    \textbf{Transformer-TPP}
    & \makecell{\textbf{0.856}\\\textbf{(0.028)}}
    & \makecell{0.710\\(0.020)}
    & \makecell{0.886\\(0.030)} \\

    \textbf{Transformer-SurvivEHR}
    & \makecell{0.756\\(0.032)}
    & \makecell{0.667\\(0.019)}
    & \makecell{0.832\\(0.044)} \\

    \textbf{Transformer-ICareFM}
    & \makecell{0.757\\(0.035)}
    & \makecell{0.641\\(0.025)}
    & \makecell{0.809\\(0.042)} \\

    \textbf{Transformer-ORA}
    & \makecell{0.833\\(0.030)}
    & \makecell{\textbf{0.723}\\\textbf{(0.020)}}
    & \makecell{\textbf{0.890}\\\textbf{(0.027)}} \\

    \midrule\midrule
    \textbf{Mamba-NTP}          & \makecell{0.808 \\ (0.029)} & \makecell{0.673 \\ (0.018)} & \makecell{0.764 \\ (0.062)} \\

    \textbf{Mamba-TPP}
    & \makecell{0.851\\(0.028)}
    & \makecell{0.698\\(0.020)}
    & \makecell{0.819\\(0.042)} \\

    \textbf{Mamba-SurvivEHR}
    & \makecell{0.779\\(0.028)}
    & \makecell{0.648\\(0.018)}
    & \makecell{0.778\\(0.044)} \\

    \textbf{Mamba-ICareFM}
    & \makecell{0.776\\(0.028)}
    & \makecell{0.668\\(0.019)}
    & \makecell{0.825\\(0.034)} \\

    \textbf{Mamba-ORA}
    & \makecell{\textbf{0.856}\\\textbf{(0.032)}}
    & \makecell{0.664\\(0.020)}
    & \makecell{0.849\\(0.034)} \\

    \bottomrule
  \end{tabular}
\end{table}

\newpage
\subsubsection{Regression}

\begin{table}[!ht]
  \centering
  \caption{$R^2$ of lab test regression for MIMIC-IV patients. \textit{Larger coefficient reflects a larger proportion of variance explained.}}
  \label{tab:r2_merged_no_bili}
  
  \begin{tabular}{lcccc}
    \toprule
    \textbf{Model} & \textbf{Creatinine} & \textbf{PaO2} & \textbf{Platelets} & \textbf{Glucose} \\
    \midrule
    \textbf{CC-Llama}            & \makecell{0.319\\(0.018)} & \makecell{0.219\\(0.005)} & \makecell{0.290\\(0.004)} &
     \makecell{0.155\\(0.015)}  \\
    \textbf{CC-Mamba}            & \makecell{0.307\\(0.017)} & \makecell{0.218\\(0.005)} & \makecell{0.261\\(0.004)} &
    \makecell{0.148\\(0.014)} \\
    \textbf{MOTOR}               & \makecell{0.556\\(0.033)} & \makecell{0.231\\(0.005)} & \makecell{0.298\\(0.004)} &
     \makecell{0.163\\(0.016)} \\
    \textbf{Most Recent}         & \makecell{\textbf{0.877}\\\textbf{(0.034)}} & \makecell{-0.431\\(0.024)} & \makecell{\textbf{0.910}\\\textbf{(0.002)}} & \makecell{-0.030\\(0.035)}\\
    \midrule\midrule % between baselines and Transformer
    \textbf{Transformer-NTP}     & \makecell{0.489\\(0.025)} & \makecell{0.259\\(0.005)} & \makecell{0.312\\(0.004)} & \makecell{0.180\\(0.017)} \\
    \textbf{Transformer-TPP}     & \makecell{0.603\\(0.030)} & \makecell{0.234\\(0.005)} & \makecell{0.310\\(0.003)} & \makecell{0.154\\(0.015)} \\
    \textbf{Transformer-SurvivEHR}               & \makecell{0.607\\(0.033)} & \makecell{0.185\\(0.004)} & \makecell{0.437\\(0.004)} &
    \makecell{0.196\\(0.019)} \\
    \textbf{Transformer-ICareFM}               & \makecell{0.784\\(0.044)} & \makecell{0.284\\(0.006)} & \makecell{0.844\\(0.002)} &
    \makecell{0.134\\(0.019)} \\
    \textbf{Transformer-ORA}     & \makecell{0.603\\(0.028)} & \makecell{0.267\\(0.005)} & \makecell{0.605\\(0.003)} & 
    \makecell{0.186\\(0.018)} \\
    \midrule\midrule % between Transformer and Mamba
    \textbf{Mamba-NTP}           & \makecell{0.595\\(0.029)} & \makecell{0.270\\(0.005)} & \makecell{0.407\\(0.003)} & \textbf{\makecell{0.202\\(0.019)}} \\
    \textbf{Mamba-TPP}           & \makecell{0.617\\(0.030)} & \makecell{0.241\\(0.005)} & \makecell{0.400\\(0.004)}  & \makecell{0.160\\(0.016)} \\

    \textbf{Mamba-SurvivEHR}     & \makecell{0.693\\(0.029)} & \makecell{0.182\\(0.004)} & \makecell{0.705\\(0.003)} & \makecell{0.190\\(0.019)} \\

    \textbf{Mamba-ICareFM}       & \makecell{0.783\\(0.046)} & \makecell{0.282\\(0.006)} & \makecell{0.855\\(0.003)} & \makecell{0.154\\(0.015)} \\
    \textbf{Mamba-ORA}           & \makecell{0.645\\(0.034)} & \makecell{\textbf{0.286}\\\textbf{(0.006)}} & \makecell{0.659\\(0.003)} &  \makecell{0.166\\(0.015)} \\
    \bottomrule
  \end{tabular}
\end{table}

\begin{table}[!ht]
  \centering
  \caption{RMSE of lab test regression for MIMIC-IV patients. \textit{Lower is better.}}
  \label{tab:rmse_merged_no_bili}
  
  \begin{tabular}{lcccc}
    \toprule
    \textbf{Model} & \textbf{Creatinine} & \textbf{PaO2} & \textbf{Platelets} & \textbf{Glucose} \\
    \midrule
    \textbf{Baseline}         & \makecell{\textbf{0.503}\\\textbf{(0.084)}} & \makecell{77.727\\(0.535)} & \makecell{\textbf{38.632}\\\textbf{(0.326)}} & \makecell{69.773\\(3.440)} \\
    \textbf{CC-Llama}         & \makecell{1.244\\(0.053)} & \makecell{57.369\\(0.466)} & \makecell{108.897\\(0.653)} & \makecell{62.800\\(3.797)} \\
    \textbf{CC-Mamba}         & \makecell{1.255\\(0.052)} & \makecell{57.421\\(0.462)} & \makecell{111.113\\(0.635)} & \makecell{63.067\\(3.774)} \\
    \textbf{MOTOR}            & \makecell{0.968\\(0.067)} & \makecell{56.920\\(0.447)} & \makecell{108.285\\(0.630)} & \makecell{62.684\\(3.672)} \\
    \midrule\midrule
    \textbf{Transformer-NTP}  & \makecell{1.038\\(0.058)} & \makecell{55.888\\(0.461)} & \makecell{107.198\\(0.647)} & \makecell{62.017\\(3.684)} \\
    \textbf{Transformer-TPP}  & \makecell{0.915\\(0.063)} & \makecell{56.816\\(0.452)} & \makecell{107.359\\(0.605)} & \makecell{63.011\\(3.641)} \\
    \textbf{Transformer-SurvivEHR} & \makecell{0.911\\(0.067)} & \makecell{58.621\\(0.467)} & \makecell{96.957\\(0.584)} & \makecell{61.416\\(3.713)} \\
    \textbf{Transformer-ICareFM} & \makecell{0.673\\(0.090)} & \makecell{54.937\\(0.430)} & \makecell{51.088\\(0.529)} & \makecell{63.730\\(3.543)} \\
    \textbf{Transformer-ORA}  & \makecell{0.915\\(0.062)} & \makecell{55.576\\(0.439)} & \makecell{81.202\\(0.593)} & \makecell{61.753\\(3.700)} \\
    \midrule\midrule
    \textbf{Mamba-NTP}        & \makecell{0.925\\(0.062)} & \makecell{55.475\\(0.451)} & \makecell{99.533\\(0.632)} & \makecell{\textbf{61.199}\\\textbf{(3.721)}} \\
    \textbf{Mamba-TPP}        & \makecell{0.899\\(0.064)} & \makecell{56.549\\(0.447)} & \makecell{100.106\\(0.567)} & \makecell{62.764\\(3.671)} \\
    \textbf{Mamba-ORA}        & \makecell{0.866\\(0.068)} & \makecell{\textbf{54.877}\\\textbf{(0.429)}} & \makecell{75.476\\(0.581)} & \makecell{62.572\\(3.647)} \\
    \bottomrule
  \end{tabular}
\end{table}

\begin{table}[!ht]
  \centering
  \caption{MAE of lab test regression for MIMIC-IV patients. \textit{Lower is better.}}
  \label{tab:mae_merged_no_bili}
  
  \begin{tabular}{lcccc}
    \toprule
    \textbf{Model} & \textbf{Creatinine} & \textbf{PaO2} & \textbf{Platelets} & \textbf{Glucose} \\
    \midrule
    \textbf{Baseline}         & \makecell{\textbf{0.193}\\\textbf{(0.002)}} & \makecell{45.537\\(0.324)} & \makecell{\textbf{25.184}\\\textbf{(0.136)}} & \makecell{33.951\\(0.272)} \\
    \textbf{CC-Llama}         & \makecell{0.803\\(0.004)} & \makecell{38.620\\(0.230)} & \makecell{78.139\\(0.312)} & \makecell{34.944\\(0.251)} \\
    \textbf{CC-Mamba}         & \makecell{0.811\\(0.004)} & \makecell{38.593\\(0.219)} & \makecell{80.145\\(0.317)} & \makecell{35.188\\(0.255)} \\
    \textbf{MOTOR}            & \makecell{0.542\\(0.004)} & \makecell{38.211\\(0.206)} & \makecell{77.215\\(0.322)} & \makecell{34.396\\(0.205)} \\
    \midrule\midrule
    \textbf{Transformer-NTP}  & \makecell{0.651\\(0.004)} & \makecell{37.102\\(0.218)} & \makecell{76.247\\(0.310)} & \makecell{34.080\\(0.200)} \\
    \textbf{Transformer-TPP}  & \makecell{0.507\\(0.003)} & \makecell{38.114\\(0.217)} & \makecell{76.565\\(0.310)} & \makecell{34.745\\(0.212)} \\
    \textbf{Transformer-SurvivEHR} & \makecell{0.522\\(0.003)} & \makecell{38.938\\(0.214)} & \makecell{69.180\\(0.279)} & \makecell{\textbf{33.071}\\\textbf{(0.190)}} \\
    \textbf{Transformer-ICareFM} & \makecell{0.350\\(0.003)} & \makecell{36.794\\(0.198)} & \makecell{34.963\\(0.182)} & \makecell{36.130\\(0.209)} \\
    \textbf{Transformer-ORA}  & \makecell{0.504\\(0.003)} & \makecell{36.859\\(0.209)} & \makecell{56.163\\(0.249)} & \makecell{33.630\\(0.204)} \\
    \midrule\midrule
    \textbf{Mamba-NTP}        & \makecell{0.547\\(0.003)} & \makecell{36.983\\(0.209)} & \makecell{70.657\\(0.284)} & \makecell{33.184\\(0.194)} \\
    \textbf{Mamba-TPP}        & \makecell{0.492\\(0.003)} & \makecell{37.667\\(0.207)} & \makecell{70.825\\(0.277)} & \makecell{34.357\\(0.202)} \\
    \textbf{Mamba-ORA}        & \makecell{0.448\\(0.003)} & \makecell{\textbf{36.319}\\\textbf{(0.199)}} & \makecell{51.451\\(0.236)} & \makecell{34.393\\(0.202)} \\
    \bottomrule
  \end{tabular}
\end{table}

\clearpage
\subsection{\cumc}
\label{app:results:cumc}
As external validation, this section shows similar results across tasks and backbones on the \cumc{} dataset, a large urban center.

\subsubsection{Classification}
\begin{table}[!ht]
  \centering
  \caption{AUROC of binary classification tasks for \cumc~patients. \textit{Larger is better.}}
  \begin{adjustbox}{max width=\textwidth}
  \begin{tabular}{lccccccc}
    \toprule
    \textbf{Model} & \textbf{Readmission} & \textbf{LOS} & \textbf{Mortality} & \textbf{AMI} & \textbf{MASLD} & \textbf{Stroke} & \textbf{Celiac} \\
    \midrule

    \textbf{XGBoost} & \makecell{0.732\\(0.003)} & \makecell{0.773\\(0.002)} & \makecell{0.875\\(0.005)} & \makecell{0.833\\(0.008)} & \makecell{0.681\\(0.009)} & \makecell{0.872\\(0.006)} & \makecell{0.650\\(0.031)} \\

    \textbf{Llama} & \makecell{0.758\\(0.003)} & \makecell{0.798\\(0.002)} & \makecell{0.892\\(0.004)} & \makecell{0.821\\(0.009)} & \makecell{0.731\\(0.008)} & \makecell{0.846\\(0.007)} & \makecell{0.604\\(0.040)} \\
    \textbf{Mamba} & \makecell{0.757\\(0.003)} & \makecell{0.784\\(0.002)} & \makecell{0.883\\(0.004)} & \makecell{0.816\\(0.009)} & \makecell{0.706\\(0.008)} & \makecell{0.865\\(0.007)} & \makecell{0.635\\(0.038)} \\

    \textbf{MOTOR} & \makecell{0.783\\(0.003)} & \makecell{0.850\\(0.002)} & \makecell{0.963\\(0.002)} & \makecell{0.853\\(0.007)} & \makecell{0.727\\(0.007)} & \makecell{0.869\\(0.006)} & \makecell{0.647\\(0.030)} \\

    \midrule \midrule% line between (MOTOR/Mamba rows) and Transformer rows

    \textbf{Transformer-NTP} & \makecell{0.760\\(0.003)} & \makecell{0.842\\(0.002)} & \makecell{0.938\\(0.003)} & \makecell{0.821\\(0.008)} & \makecell{0.713\\(0.009)} & \makecell{0.849\\(0.007)} & \makecell{0.624\\(0.039)} \\

    \textbf{Transformer-TPP} & \makecell{0.784\\(0.003)} & \makecell{0.857\\(0.002)} & \makecell{0.963\\(0.002)} & \makecell{0.843\\(0.008)} & \makecell{0.718\\(0.008)} & \makecell{0.863\\(0.006)} & \makecell{0.736\\(0.033)} \\

    \textbf{Transformer-SurvivEHR} & \makecell{0.749\\(0.003)} & \makecell{0.795\\(0.002)} & \makecell{0.923\\(0.004)} & \makecell{0.826\\(0.008)} & \makecell{0.709\\(0.008)} & \makecell{0.839\\(0.006)} & \makecell{0.629\\(0.031)} \\

    \textbf{Transformer-ICareFM} & \makecell{0.760\\(0.003)} & \makecell{0.846\\(0.002)} & \makecell{0.948\\(0.003)} & \makecell{0.825\\(0.008)} & \makecell{0.702\\(0.008)} & \makecell{0.828\\(0.007)} & \makecell{0.594\\(0.035)} \\

    \textbf{Transformer-ORA} & \textbf{\makecell{0.787\\(0.003)}} & \textbf{\makecell{0.866\\(0.002)}} & \textbf{\makecell{0.968\\(0.002)}} & \makecell{0.850\\(0.008)} & \makecell{0.745\\(0.007)} & \makecell{0.870\\(0.007)} & \textbf{\makecell{0.749\\(0.036)}} \\

    \midrule\midrule
    \textbf{Mamba-NTP} & \makecell{0.757\\(0.003)} & \makecell{0.817 \\ (0.002)} & \makecell{0.915\\(0.003)} & \makecell{0.780\\(0.010)} & \makecell{0.681\\(0.008)} & \makecell{0.830\\(0.007)} & \makecell{0.637\\(0.033)} \\
    \textbf{Mamba-TPP} & \makecell{0.776\\(0.003)} & \makecell{0.859\\(0.002)} & \makecell{0.942\\(0.003)} & \makecell{0.854\\(0.008)} & \makecell{0.737\\(0.008)} & \makecell{0.864\\(0.006)} & \makecell{0.739\\(0.034)} \\
    \textbf{Mamba-ORA} & \makecell{0.779\\(0.002)} & \makecell{0.861\\(0.002)} & \makecell{0.946\\(0.003)} & \textbf{\makecell{0.858\\(0.008)}} & \textbf{\makecell{0.766\\(0.008)}} & \textbf{\makecell{0.873\\(0.006)}} & \makecell{0.744\\(0.038)} \\
    \bottomrule
  \end{tabular}
  \end{adjustbox}
\end{table}

\newpage
\subsubsection{Time-to-event}

\begin{table}[!ht]
  \centering
  \caption{C-index of time-to-event prediction tasks for \cumc~patients. \textit{Larger is better.}}

  \begin{tabular}{lcccc}
    \toprule
    \textbf{Model} & \textbf{AMI} & \textbf{Celiac} & \textbf{Stroke} & \textbf{MASLD} \\
    \midrule

    \textbf{Deephit} 
    & \makecell{0.615\\(0.008)} 
    & \makecell{0.495\\(0.029)} 
    & \makecell{0.637\\(0.008)} 
    & \makecell{0.602\\(0.006)} \\ 
    
    \textbf{CC-Llama}
    & \makecell{0.758\\(0.011)} 
    & \makecell{0.622\\(0.034)} 
    & \makecell{0.808\\(0.009)} 
    & \makecell{0.704\\(0.009)} \\ 
    \textbf{CC-Mamba} 
    & \makecell{0.774\\(0.011)} 
    & \makecell{0.693\\(0.036)} 
    & \makecell{0.815\\(0.009)} 
    & \makecell{0.692\\(0.009)} \\ 
    
    \textbf{MOTOR} 
    & \makecell{0.792\\(0.012)} 
    & \makecell{0.645\\(0.036)} 
    & \makecell{0.804\\(0.009)} 
    & \makecell{0.674\\(0.009)} \\

    \midrule\midrule

    \textbf{Transformer-NTP} 
    & \makecell{0.724\\(0.012)} 
    & \makecell{0.564\\(0.035)} 
    & \makecell{0.769\\(0.010)} 
    & \makecell{0.637\\(0.009)} \\

    \textbf{Transformer-TPP} 
    & \makecell{0.775\\(0.014)} 
    & \makecell{0.694\\(0.031)} 
    & \makecell{0.795\\(0.010)} 
    & \makecell{0.664\\(0.010)} \\

    \textbf{Transformer-SurvivEHR} 
    & \textbf{\makecell{0.821\\(0.009)}} 
    & \makecell{0.681\\(0.033)} 
    & \makecell{0.810\\(0.010)} 
    & \makecell{0.700\\(0.009)} \\

    \textbf{Transformer-ICareFM} 
    & \makecell{0.767\\(0.009)} 
    & \makecell{0.667\\(0.028)} 
    & \makecell{0.742\\(0.011)} 
    & \makecell{0.667\\(0.010)} \\

    \textbf{Transformer-ORA} 
    & \makecell{0.792\\(0.013)} 
    & \textbf{\makecell{0.743\\(0.029)}} 
    & \makecell{0.793\\(0.010)} 
    & \makecell{0.692\\(0.011)} \\

    \midrule\midrule

    \textbf{Mamba-NTP} 
    & \makecell{0.730\\(0.011)} 
    & \makecell{0.507\\(0.043)} 
    & \makecell{0.782\\(0.009)} 
    & \makecell{0.669\\(0.008)} \\

    \textbf{Mamba-TPP} 
    & \makecell{0.767\\(0.012)} 
    & \makecell{0.651\\(0.034)} 
    & \textbf{\makecell{0.823\\(0.010)}} 
    & \textbf{\makecell{0.708\\(0.010)}} \\

    \textbf{Mamba-ORA} 
    & \makecell{0.785\\(0.012)} 
    & \makecell{0.725\\(0.038)} 
    & \makecell{0.788\\(0.011)} 
    & \makecell{0.704\\(0.009)} \\

    \bottomrule
  \end{tabular}
\end{table}

\newpage
\subsubsection{Regression}

\begin{table}[!ht]
  \centering
  \caption{$R^2$ of lab test regression for \cumc~patients. \textit{Larger coefficient reflects a larger proportion of variance explained.}}
  \label{tab:r2_transformer_only_no_bili}
  
  \begin{tabular}{lcccc}
    \toprule
    \textbf{Model} & \textbf{Creatinine} & \textbf{Platelets} & \textbf{SpO2} & \textbf{Glucose} \\
    \midrule
    \textbf{Most Recent}        & \makecell{0.679\\(0.036)} & \makecell{0.426\\(0.011)} & \makecell{0.078\\(0.013)} & \makecell{0.052 \\ (0.015)} \\
    \textbf{CC-Llama}           & \makecell{0.298\\(0.013)} & \makecell{0.150\\(0.005)} & \makecell{0.770\\(0.007)} & \makecell{0.217 \\(0.007)} \\
    \textbf{CC-Mamba}           & \makecell{0.277\\(0.013)} & \makecell{0.146\\(0.006)} & \makecell{0.762\\(0.007)} & \makecell{0.196 \\(0.006)} \\
    \textbf{MOTOR}              & \makecell{0.469\\(0.020)} & \makecell{0.198\\(0.007)} & \makecell{0.842\\(0.006)} & \makecell{0.282\\(0.008)} \\
    \midrule \midrule
    \textbf{Transformer-NTP}    & \makecell{0.431\\(0.022)} & \makecell{0.228\\(0.006)} & \makecell{0.816\\(0.007)} & \makecell{0.287\\(0.009)} \\
    \textbf{Transformer-TPP}    & \makecell{0.397\\(0.018)} & \makecell{0.203\\(0.007)} & \makecell{0.832\\(0.006)} & \makecell{0.279\\(0.008)} \\
    \textbf{Transformer-SurvivEHR} & \makecell{0.534\\(0.024)} & \makecell{0.481\\(0.007)} & \makecell{0.798\\(0.006)} & \makecell{0.365\\(0.009)} \\
    \textbf{Transformer-ICareFM} & \makecell{\textbf{0.708}\\\textbf{(0.031)}} & \makecell{\textbf{0.568}\\\textbf{(0.008)}} & \makecell{0.846\\(0.006)} & \makecell{\textbf{0.431}\\\textbf{(0.009)}} \\
    \textbf{Transformer-ORA}    & \makecell{0.542\\(0.024)} & \makecell{0.349\\(0.008)} & \makecell{0.836\\(0.006)} & \makecell{0.305\\(0.008)} \\
    \midrule\midrule
    \textbf{Mamba-NTP}          & \makecell{0.423\\(0.024)} & \makecell{0.220\\(0.007)} & \makecell{0.809\\(0.006)} & \makecell{0.208\\(0.007)} \\
    \textbf{Mamba-TPP}          & \makecell{0.543\\(0.027)} & \makecell{0.311\\(0.008)} & \makecell{0.849\\(0.006)} & \makecell{0.306\\(0.008)} \\
    \textbf{Mamba-ORA}          & \makecell{0.582\\(0.029)} & \makecell{0.464\\(0.007)} & \makecell{\textbf{0.852}\\\textbf{(0.006)}} & \makecell{0.330\\(0.008)} \\
    \bottomrule
  \end{tabular}
\end{table}

\begin{table}[!ht]
  \centering
  \caption{RMSE of lab test regression for \cumc~patients. \textit{Lower is better.}}
  \label{tab:rmse_transformer_only_no_bili}
  
  \begin{tabular}{lcccc}
    \toprule
    \textbf{Model} & \textbf{Creatinine} & \textbf{Platelets} & \textbf{SpO2} & \textbf{Glucose} \\
    \midrule
    \textbf{Most Recent}        & \makecell{0.630\\(0.047)} & \makecell{74.419\\(1.048)} & \makecell{15.598\\(0.225)} & \makecell{68.571 \\ (0.78)} \\
    \textbf{CC-Llama}           & \makecell{0.987\\(0.035)} & \makecell{90.486\\(0.941)} & \makecell{7.802\\(0.114)} & \makecell{62.172 \\ (0.703)} \\
    \textbf{CC-Mamba}           & \makecell{1.001\\(0.035)} & \makecell{90.695\\(0.939)} & \makecell{7.939\\(0.112)} & \makecell{63.002 \\(0.742)}  \\
    \textbf{MOTOR}              & \makecell{0.825\\(0.042)} & \makecell{88.308\\(0.948)} & \makecell{6.437\\(0.123)} & \makecell{59.754\\(0.624)} \\
    \midrule \midrule
    \textbf{Transformer-NTP}    & \makecell{0.856\\(0.039)} & \makecell{86.837\\(0.969)} & \makecell{6.948\\(0.110)} & \makecell{59.545\\ (0.580)} \\
    \textbf{Transformer-TPP}    & \makecell{0.879\\(0.041)} & \makecell{88.046\\(0.936)} & \makecell{6.638\\(0.117)} & \makecell{59.888\\(0.616)} \\
    \textbf{Transformer-SurvivEHR} & \makecell{0.772\\(0.043)} & \makecell{71.032\\(0.861)} & \makecell{7.289\\(0.117)} & \makecell{56.187\\(0.590)} \\
    \textbf{Transformer-ICareFM} & \makecell{\textbf{0.612}\\\textbf{(0.049)}} & \makecell{\textbf{64.829}\\\textbf{(0.838)}} & \makecell{6.354\\(0.122)} & \makecell{\textbf{53.214}\\\textbf{(0.546)}} \\
    \textbf{Transformer-ORA}    & \makecell{0.766\\(0.044)} & \makecell{79.574\\(0.883)} & \makecell{6.555\\(0.122)} & \makecell{58.804\\(0.609)} \\
    \midrule \midrule
    \textbf{Mamba-NTP}          & \makecell{0.860\\(0.045)} & \makecell{87.097\\(0.881)} & \makecell{7.078\\(0.110)} & \makecell{62.749\\(0.663)} \\
    \textbf{Mamba-TPP}          & \makecell{0.765\\(0.047)} & \makecell{81.869\\(0.911)} & \makecell{6.301\\(0.123)} & \makecell{58.752\\(0.609)} \\
    \textbf{Mamba-ORA}          & \makecell{0.732\\(0.047)} & \makecell{72.188\\(0.878)} & \makecell{\textbf{6.226}\\\textbf{(0.118)}} & \makecell{57.703\\(0.592)} \\
    \bottomrule
  \end{tabular}
\end{table}

\begin{table}[!ht]
  \centering
  \caption{MAE of lab test regression for \cumc~patients. \textit{Lower is better.}}
  \label{tab:mae_transformer_only_no_bili}
  
  \begin{tabular}{lcccc}
    \toprule
    \textbf{Model} & \textbf{Creatinine} & \textbf{Platelets} & \textbf{SpO2} & \textbf{Glucose} \\
    \midrule
    \textbf{Most Recent}        & \makecell{\textbf{0.235}\\\textbf{(0.004)}} & \makecell{47.591\\(0.428)} & \makecell{6.377\\(0.112)} & \makecell{45.428 \\(0.36)} \\
    \textbf{CC-Llama}           & \makecell{0.567\\(0.006)} & \makecell{65.073\\(0.458)} & \makecell{4.670\\(0.046)} & \makecell{42.073 \\(0.322)} \\
    \textbf{CC-Mamba}           & \makecell{0.567\\(0.006)} & \makecell{65.327\\(0.488)} & \makecell{4.742\\(0.045)} & \makecell{42.848 \\ (0.350)} \\
    \textbf{MOTOR}              & \makecell{0.402\\(0.006)} & \makecell{63.290\\(0.478)} & \makecell{3.325\\(0.038)} & \makecell{39.864\\(0.289)} \\
   \midrule\midrule
    \textbf{Transformer-NTP}    & \makecell{0.445\\(0.005)} & \makecell{62.540\\(0.436)} & \makecell{3.968\\(0.037)} & \makecell{40.257\\ (0.299)} \\
    \textbf{Transformer-TPP}    & \makecell{0.406\\(0.006)} & \makecell{63.068\\(0.435)} & \makecell{3.448\\(0.037)} & \makecell{39.893\\(0.274)} \\
    \textbf{Transformer-SurvivEHR} & \makecell{0.367\\(0.005)} & \makecell{48.594\\(0.339)} & \makecell{4.145\\(0.045)} & \makecell{36.864\\(0.267)} \\
    \textbf{Transformer-ICareFM} & \makecell{0.275\\(0.004)} & \makecell{\textbf{43.674}\\\textbf{(0.323)}} & \makecell{3.253\\(0.040)} & \makecell{\textbf{35.178}\\\textbf{(0.249)}} \\
    \textbf{Transformer-ORA}    & \makecell{0.341\\(0.005)} & \makecell{56.882\\(0.376)} & \makecell{3.377\\(0.040)} & \makecell{38.962\\(0.287)} \\
    \midrule\midrule
    \textbf{Mamba-NTP}          & \makecell{0.434\\(0.006)} & \makecell{62.555\\(0.393)} & \makecell{4.086\\(0.039)} & \makecell{42.080\\(0.286)} \\
    \textbf{Mamba-TPP}          & \makecell{0.340\\(0.005)} & \makecell{58.465\\(0.407)} & \makecell{3.196\\(0.038)} & \makecell{38.868\\(0.282)} \\
    \textbf{Mamba-ORA}          & \makecell{0.326\\(0.005)} & \makecell{49.793\\(0.357)} & \makecell{\textbf{3.103}\\\textbf{(0.037)}} & \makecell{37.976\\(0.276)} \\
    \bottomrule
  \end{tabular}
\end{table}

\FloatBarrier
\newpage

\end{document}